\documentclass[10pt, a4paper]{article}
\usepackage{lrec-coling2024} 
\usepackage{times}
\usepackage{latexsym}

\usepackage{graphicx}
\usepackage{amsmath}
\usepackage{amsmath}
\usepackage{subfig}
\usepackage{array, makecell} 
\usepackage{multirow}
\usepackage{todonotes}
\usepackage{paralist, tabularx}
\usepackage{amsmath}
\usepackage{ amssymb }
\usepackage{ mathrsfs }
\usepackage{float}
\usepackage{hyperref}
\usepackage[T1]{fontenc}
\usepackage{ comment }

\title{Labeling Comic Mischief Content in Online Videos with a Multimodal Hierarchical-Cross-Attention Model}

\name{Elaheh Baharlouei$^{1}$, Mahsa Shafaei$^{1}$, Yigeng Zhang$^{1}$
\\ \bf \large{Hugo Jair Escalante$^{2}$, Thamar Solorio$^{1,3}$}}

\address{$^{1}$ Department of Computer Science, University of Houston, Houston, USA\\
\{ebaharlouei, mshafaei, yzhang168, tsolorio\}@uh.edu\\
$^{2}$ Department of Computer Science, INAOE, Puebla, Mexico \\
hugojair@inaoep.mx\\
$^{3}$ MBZUAI, Masdar City, Abu Dhabi, UAE}

\abstract{
We address the challenge of detecting questionable content in online media, specifically the subcategory of comic mischief. This type of content combines elements such as violence, adult content, or sarcasm with humor, making it difficult to detect. Employing a multimodal approach is vital to capture the subtle details inherent in comic mischief content.
To tackle this problem, we propose a novel end-to-end multimodal system for the task of comic mischief detection. 
As part of this contribution, we release a novel dataset for the targeted task consisting of three modalities: video, text (video captions and subtitles), and audio. We also design a HIerarchical Cross-attention model with CAPtions (HICCAP) to capture the intricate relationships among these modalities. 
The results show that the proposed approach makes a significant improvement over robust baselines and state-of-the-art models for comic mischief detection and its type classification. This emphasizes the potential of our system to empower users, to make informed decisions about the online content they choose to see.
In addition, we conduct experiments on the UCF101, HMDB51, and XD-Violence datasets, comparing our model against other state-of-the-art approaches showcasing the outstanding performance of our proposed model in various scenarios.
 \\ \newline  }

\begin{document}

\maketitleabstract

\section{Introduction}
\label{sec:intro}
\vspace{-0.2cm}
The impact of media on children has long been debated in psychology~\cite {dietz1991children}. Regardless of the positive impacts of media on children's education, multiple studies demonstrate that violent and aggressive material negatively affects children's behavior \cite{wilson2008media,chang2019effect}. Online media may have detrimental effects beyond violence, for example, by normalizing negative behaviors such as drug and alcohol abuse or premature engagement in sexual activities in society ~\cite{hanewinkel2014portrayal,strasburger1989adolescent}.

Automated systems for categorizing online content based on the presence of questionable material have the potential to make a significant difference in protecting users from distressing and possibly objectionable content. Identifying content with intuitive labels offers a more flexible alternative to the commonly used age-based rating systems, such as those by the Motion Picture Association of America. This flexibility accounts for the fact that people's tolerance for questionable content can vary widely, influenced by factors like age, life experiences, socio-cultural values, and cognitive skills \cite{anderson2003influence}.
In this paper, for the first time, we focus on detecting comic mischief content in videos, which is a subset of questionable content. In a comic mischief video, questionable content (violence, adult content, or sarcastic material) is combined with a humorous context, making it even more disruptive. According to psychologists, when something such as violence is presented in a serious context (such as war), it has a less disruptive effect than when it is presented in a pleasant and humorous context \cite{blackford2011prevalence}.

Automated detection of comic mischief in videos poses significant challenges. The difficulty lies in differentiating between serious content and humor, particularly when the humor involves subtle jokes or culturally specific knowledge. People's perceptions of humor can vary greatly due to cultural backgrounds and personal preferences, complicating the establishment of universal criteria for humor detection. Moreover, videos comprise visual imagery, audio, and text, necessitating a comprehensive approach that considers all these elements \cite{yang2023multi}.
In response to these challenges, our paper introduces a multimodal approach designed to capture the diverse information presented among the modalities to better understand and identify comic mischief. We present HICCAP, an innovative multimodal system specifically developed for detecting comic mischief. It features a hierarchical cross-attention module that identifies intermodal relationships, enhancing our system's effectiveness.
Our approach includes a binary model for determining the presence of comic mischief and a multi-task model that categorizes different types of comic mischief. The models analyze three key modalities found in our dataset: visual content, audio signals, and textual information from dialogues.
The proposed model is pretrained using a multimodal hybrid pretraining technique that integrates both matching tasks and contrastive-learning methods on extensive multimodal datasets. 
Furthermore, our empirical research demonstrates that using a pretrained video captioning model effectively supplements subtitles in videos that lack them and boosts the model's performance.
Complementing our multimodal system, we introduce a new dataset for the comic mischief task, collected from a mix of freely accessible internet videos and the Youtube-8M \cite{abu2016youtube} corpus.  This collection has been meticulously annotated for comic mischief categories by twelve annotators. 
This dataset can be a valuable resource for the community by giving detailed information about the comic mischief content across multiple video modalities (i.e., dialogue, sound, and video).
While it will chiefly benefit those in natural language processing, computer vision, machine learning, and signal processing, the dataset also holds the potential to inspire new inquiries in social sciences, child psychology, and mass media fields.

The contributions of this paper are as follows:
\begin{compactitem}
    \item Introduces a new video classification task and provides a novel dataset, aiming to motivate further research on this relevant topic\footnote{\href{https://github.com/RiTUAL-UH/Comic-Mischief-Prediction}{https://github.com/RiTUAL-UH/Comic-Mischief-Prediction}}.
    
    \item Proposes a new end-to-end multimodal model based on a hierarchical cross-attention module to effectively capture the relationships between different modalities.
    \item Incorporates an automatic captioning technique to fill up the gaps in video subtitles that considerably boosts the model's performance. 

\end{compactitem}
\vspace{-0.2cm}

\section{Multimodal Comic Mischief Dataset}
\label{sec:dataset}
To create technology for labeling comic mischief in videos, we need access to a repository of consistently labeled video content. This requires collecting a diverse selection of online videos and using a customized annotation tool to implement an incremental annotation strategy. In this section, we introduce the
comic mischief dataset we are releasing with this paper.

\begin{figure}[h!]
\centering
\subfloat[Gory Humor]{
  \includegraphics[width=0.43\columnwidth]{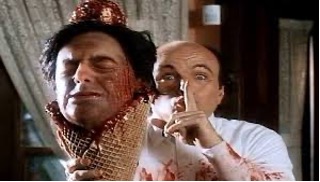}
  \label{fig:emotionsgroup-a}
}
\subfloat[Slapstick]{%
  \includegraphics[width=0.37\columnwidth]{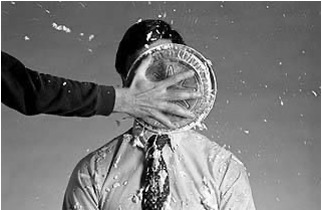}%
  \label{fig:emotionsgroup-b}
}


\subfloat[Mature Humor]{
  \includegraphics[width=0.42\columnwidth]{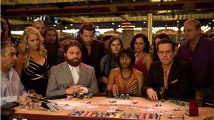}%
  \label{fig:emotionsgroup-b}
}
\subfloat[Sarcasm]{
  \includegraphics[width=0.35\columnwidth]{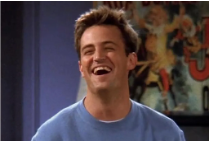}%
  \label{fig:emotionsgroup-b}
}
\caption{Comic mischief examples in movies}
\label{fig:qsamples}
\end{figure}

In our dataset, each comic mischief label may be associated with any or all three considered modalities (sound, dialogue, and video). 
The definition of each modality is as follows: \textbf{Dialogue:} transcription of spoken dialogue between characters (subtitle of video);
\textbf{Sound:} sound effects and ambient sounds (e.g., explosions);
\textbf{Video:} intensity (the amount of light or the numerical value of a pixel) information from video frames.
We considered four categories of comic mischief including: 
\begin{compactitem}
    \item \textbf{Gory Humor:} refers to a situation with a great deal of bloodshed and violence juxtaposed with humorous references. 
\item \textbf{Slapstick  Humor:} is a comedy style characterized by practical jokes, collisions, clumsiness, and embarrassing events (e.g., people get poked in the eye or pies in the face). 
\item \textbf{Mature Humor:} strong language, alcohol/drug consumption, gambling, and sexual references in depictions or dialogue with humor references. 
\item \textbf{Sarcasm:} employs words to mock or annoy someone for humorous effect. Sarcasm may employ ambiguity, but it is not always ironic. 
\end{compactitem}
Figure \ref{fig:qsamples} shows sample screenshots of comic mischief categories.

\subsection{Video Collection Process}
Our goal is to collect a wide range of English-language videos with comic mischief material. For our data collection, we leverage two primary sources.
Firstly, we gather data from YouTube, ensuring that our methods adhere to YouTube's policies. In alignment with these policies, we will release a JSON file containing relevant YouTube URLs for the dataset, making it accessible for viewing and downloading. YouTube is one of our platforms of choice given its widespread availability to internet users. We developed an algorithm that utilizes YouTube's recommendation system to discover new videos. The methodology includes:
1) Conducting an initial manual search to identify seed videos on the desired topic.
2) Generating a list of videos recommended for each seed video. Given YouTube's content-based recommendation mechanism, some content overlap is anticipated. These suggested videos then became our new seed material.
3) Repeating the above process three times to ensure diversity among videos. To maintain consistency in theme, with each iteration, we selected fewer videos per seed.
Our second resource is the YouTube-8M dataset, which is a freely available dataset for the research community. This corpus is a collection of human-verified labels on about 237K segments on 1000 classes. To ensure varied video content, our primary focus is on categories such as talk shows, sketch comedy, comedy (drama), sitcoms, and stand-up comedies.

To facilitate our research, we divided the videos into 60-second clips and proceeded to annotate these clips. In the remainder of this paper, we use the term clip while referring to the 60-second segments of a video in our dataset.
In total, we have 4478 clips extracted from 1179 videos. From the first source, YouTube, we collected 1233 clips originating from 394 videos. From the YouTube-8M dataset, our second source, we gathered 3245 clips derived from 785 videos. 

\subsection{Annotation Process}
\begin{figure}[!htpb]
\centering
    \includegraphics[width=0.8\columnwidth]{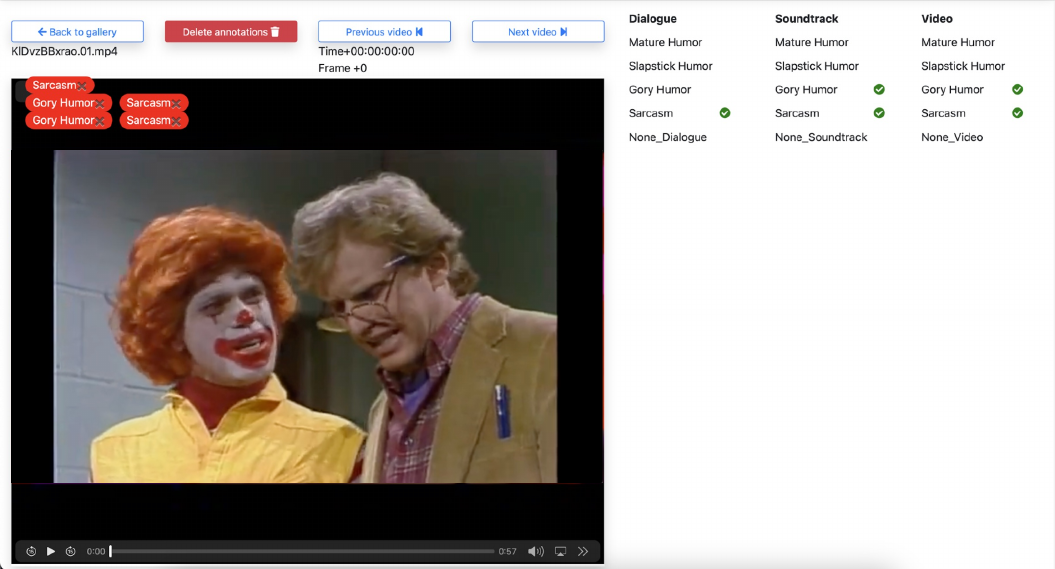}
    \caption{User interface for labeling dataset videos.} 
    \label{interface}
\end{figure}

As a first step, we must enrich the taxonomy of questionable content and establish guidelines for annotators. To create a guideline for annotation, we did the following steps: 
1) established criteria for annotation of questionable content and developed annotation guidelines; 2) compiled a list of diverse online videos for use in pilot annotations; 3) to revise the annotation process and components, we conducted several workshops with participants from the fields of psychology, AI ethics, computer vision, and natural language processing; 4) finally, we fine-tuned the guidelines in light of pilot annotation experience and discussion.
To ensure the quality of the annotation process, we have implemented a three-way web-based annotation interface displaying each clip with its audio and transcript. This interface facilitates annotations across various comic mischief categories in three modalities, with each clip being evaluated by three annotators. Figure \ref{interface} illustrates our web interface. 
Twelve annotators participated in the task, including seven graduate and five bachelor students. Of these, seven were native English speakers, while the remaining five were fluent in English.
Each label has a binary value, indicating the presence of comic mischief. Also, each label can be displayed in any of the clip's modalities (Dialogue, Sound, Video). As a result, we have distinct labels for each modality (e.g., Gory Humor - Dialogue, Gory  Humor - Video), and one clip can belong to more than one category. A majority vote of all annotators determines the final value assigned to each label per modality. 
To evaluate the annotation's quality, we have computed the Inter-Annotator Agreement (IAA) by calculating Cohen's Kappa ($\kappa$) on each annotator's annotation and the majority voting of all annotations. Based on the computed IAA values, there is a substantial agreement ($\kappa=0.70$).
\vspace{-0.1cm}
\subsection{Dataset Statistics}
\vspace{-0.1cm}
Table \ref{tabel:stsall} provides an overview of the class-level statistics for the video segments. As the table shows, some videos do not contain dialogue (e.g., silent films). Videos in class 1 (C1) contain at least one type of comic mischief, while those in class 0 (C0) do not. This table shows that each class has an equal distribution of video length and content. The statistics demonstrate that the length of videos and the average number of words per video are comparable for both classes, ensuring that they contain enough content to be meaningful. 
\begin{table}[h!]
\centering
\scalebox{0.69}{
\begin{tabular}{|l|c|c|c|c|c|c|c|c|}
\hline
            & \multicolumn{2}{c|}{Max} & \multicolumn{2}{c|}{Min}  & \multicolumn{2}{c|}{Avg}  & \multicolumn{2}{c|}{Med} \\ \hline
            & C0   & C1  & C0    & C1   & C0     & C1     & C0    & C1   \\ \hline
\# Words    & 259 & 266  & 0     & 0    & 106   & 118    & 111   & 125    \\ \hline
V/A Length & 64.9 & 71.9 & 0.1   & 9.4  & 54.7  & 58.6   & 60.1 & 60.5  \\ \hline
\# Frames  & 1836 & 2157  & 1    & 108  & 538   & 658    & 460  & 478    \\ \hline

\end{tabular}}
\caption{Video segments statistics, C0, C1, and V/A Length stand for class 0, class 1, and video or audio clip length respectively.~\label{tabel:stsall}}
\end{table}

\begin{figure}[!htpb]
\centering
    \includegraphics[width=0.65\columnwidth]{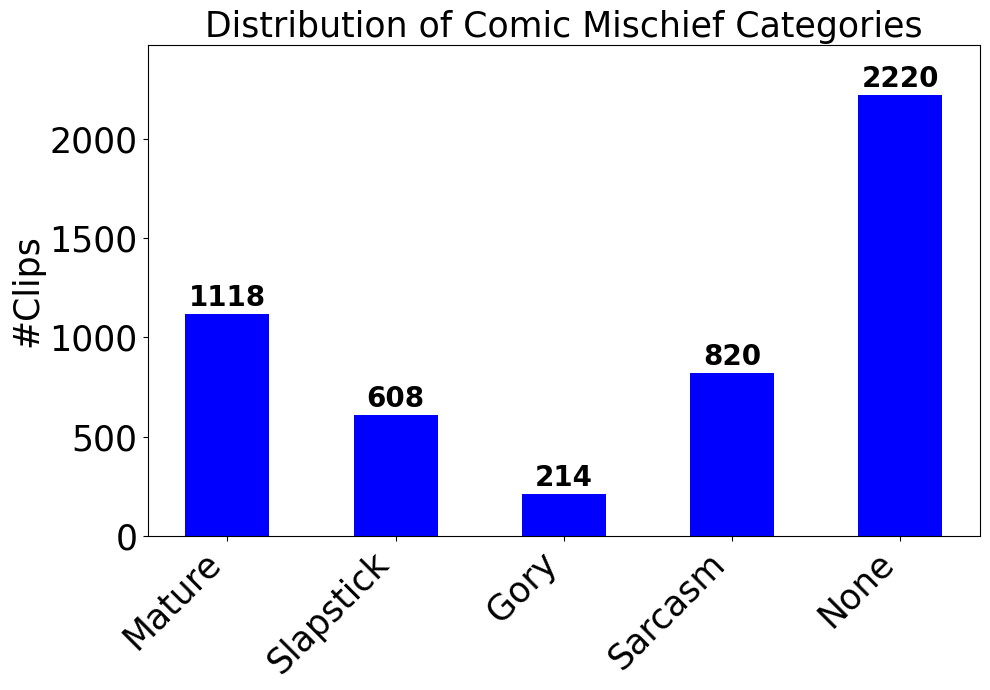}
    \caption{Distribution of comic mischief categories. 
    }
    \label{multi-class-statistic}
\end{figure}

Figure~\ref{multi-class-statistic} displays the data distribution across each category for the 
multi-label dataset. 
As shown in this figure, classes are not balanced: the majority class (Mature Humor) has more than five times the samples for the minority class (Gory Humor). 

The introduced dataset is the first one approaching the comic mischief detection task. This dataset poses a number of challenges including: video classification with imbalanced classes, the need for effective use of multimodal information for classification, understanding subtle humor cues intertwined with potentially harmful content, discerning the intricate relationships between visual, audio, and textual modalities in the context of comic mischief, and handling data where audio and text may sometimes conflict or diverge from visual content.
\vspace{-0.3cm}

\section{Methodology}
\label{sec:methodology}
In this section, we describe an automated model for categorizing videos containing comic mischief material. We approach this problem from two perspectives: first, as a binary classification task, and second, by developing a multi-task
model that can concurrently predict all four subcategories. 

\begin{figure*}[!htpb]
\centering
    \includegraphics[width=0.9\textwidth]{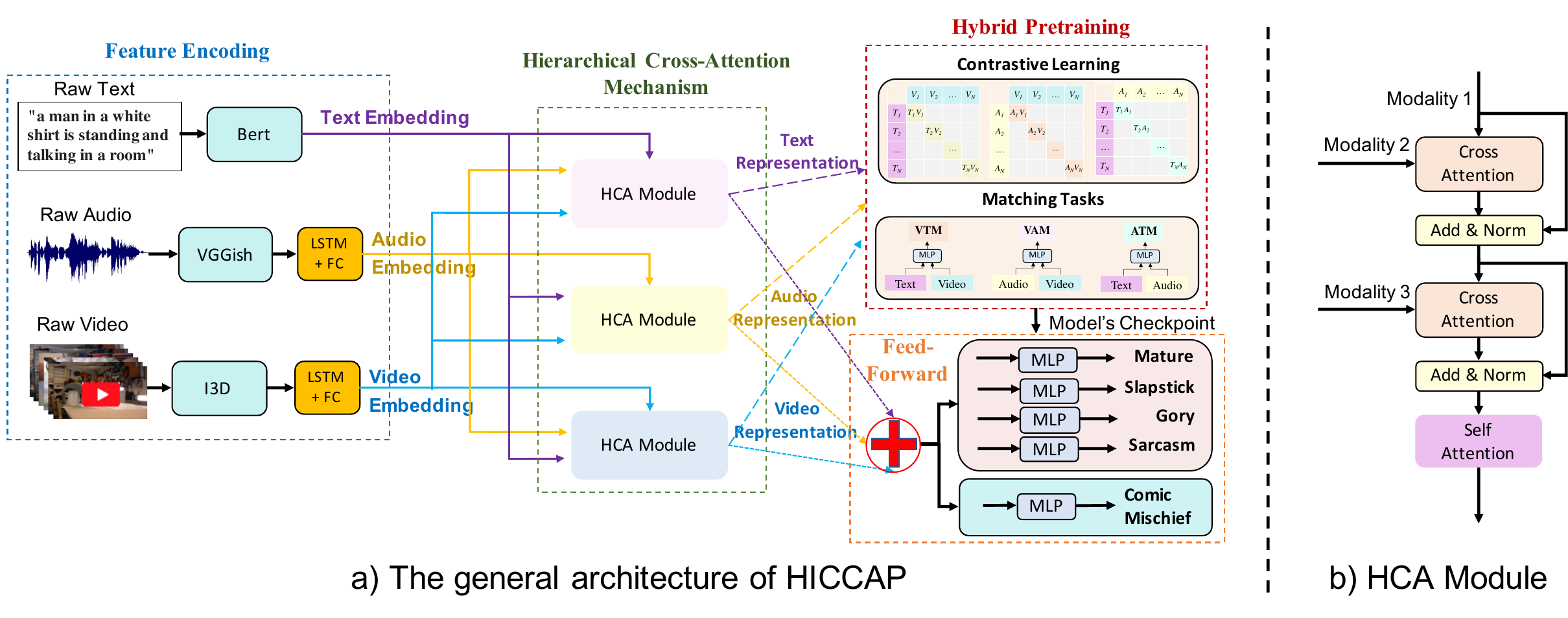}
    \caption{a) The general architecture of HICCAP consists of four components: 1) Feature-Encoding, 2) Hierarchical-Cross-Attention mechanisms, 3) Pretraining, and 4) Binary and Multi-Task Prediction and b) The structure of Hierarchical Cross-Attention (HCA) module.}
    \label{Overal-structure}
\end{figure*}

Our "HIerarchical Cross-attention model with CAPtions" (HICCAP) is a unified end-to-end model designed to capture the inherent complexities of comic mischief prediction and type classification. We encoded each modality (audio, text, and video) with pretrained models, and then fed the encoded audio and video vectors into RNNs to account for the sequential information. Recognizing that one modality can impact the interpretation or significance of others (e.g., the speaker's tone may alter the meaning of their words), we introduce a cross-attention network to capture these intermodality relationships. Figure \ref{Overal-structure} provides the system overview, with subsequent sections detailing each module.


\subsection{Feature Encoding}
To better represent a video, we utilize pretrained models to extract features from various modalities, including raw text, audio, and video, ensuring a comprehensive understanding of the context.

We use the pretrained BERT model~\cite{devlin2018bert} for encoding the input text in our model.
Since some of the videos lack dialogue, they do not include subtitles. To fill this void, we use a pre-trained model for video captioning to generate captions for these videos and use them in the absence of subtitles. We employ Dense video Captioning with a Bi-modal Transformer (BMT) for this purpose \cite{iashin2020better}. In BMT, the model locates important events in a video and generates a unique textual description for each event.


We utilize the pretrained VGGish network~\cite{hershey2017cnn} for audio feature extraction and the pretrained I3D network~\cite{carreira2017quo} for visual modality encoding. For both audio and video vectors, we process them through an LSTM to extract sequential information, followed by a Fully Connected (FC) layer.

\vspace{-0.2cm}
\subsection{Hierarchical Cross-Attention }
\vspace{-0.1cm}
In multimodal videos, the content of each modality may affect the meaning or importance of other modalities.
To capture this intermodality relation, 
we utilize the cross-attention mechanism 
to enhance the representation of multiple modalities. 
In cross-attention, the query belongs to one modality, while the key and value vectors belong to the context modality ~\cite{zheng2020cross,lu2019vilbert}.
To leverage the effectiveness of cross-attention for multiple modalities, we introduce a new hierarchical cross-attention mechanism. 

While the concept of hierarchical cross-attention has been explored in earlier studies \cite{chen2022efficient, zhang2022hierarchical, yang2023bert}, our proposed Hierarchical Cross-Attention technique, HICCAP, offers unique features and benefits. Notably, the methods presented in \cite{chen2022efficient, yang2023bert} are restricted to a single modality, and \cite{zhang2022hierarchical} focuses solely on image and text. In contrast, HICCAP encompasses three modalities: video, audio, and text, and uniquely aims to encode the significance of a single modality based on the interactions with the other modalities. To our knowledge, this is the first work that considers all these modalities together. Additionally, HICCAP introduces a distinct Hierarchical Cross-Attention (HCA) module (Figure \ref{Overal-structure}.b) for every modality and strategically combines these HCA modules, ensuring that while emphasizing one modality, the attention from the other two modalities is also considered. 


In our hierarchical cross-attention approach, the first cross-attention layer calculates the attention for modality $M_1$ using $M_2$. This output subsequently forms the query vector, while modality $M_3$ serves as both key and value vectors for the following cross-attention layer. This ensures that $M_1$'s attention is influenced by both $M_2$ and $M_3$. The procedure can be formalized as shown in Equation \ref{eq:3}, where K, Q, and V denote key, query, and value, respectively, and $d_k$ represents the key vector's dimension.
In the hierarchical cross-attention approach, the order in which modalities are processed is pivotal. The sequence dictates how information flows and integrates, influencing the final representation. Consequently, selecting an appropriate ordering for modalities, based on the characteristics of the dataset and task, can impact the model's effectiveness and understanding. We did an empirical experiment to find the best ordering of modalities in different hierarchical attention modules. 
%
\begin{equation}
\resizebox{0.7\hsize}{!}{$\begin{array}{ll}
\begin{gathered}
\label{eq:3}
 head_1= \mbox{softmax}(\frac{K^T_{m2} Q_{m1}}{\sqrt{d_k}})V_{m2} \\
 head^{m1}= \mbox{softmax}(\frac{K^T_{m3} Q_{head_1}}{\sqrt{d_k}})V_{m3} \\
 \end{gathered}
\end{array}$}
\end{equation}

In Equation~\ref{eq:3}, $head^{m1}$ shows the representation of modality $M_1$ based on modalities $M_2$ and $M_3$. We repeat this process for calculating $head^{m2}$ and $head^{m3}$ which are the representations of $M_2$ and $M_3$, respectively. Then, we pass each representation separately through the attention layer, 
\cite{bahdanau2014neural}. This layer computes the weighted sum $r$ as $\sum_i\alpha_ihead^{mj}_i$ where $j \in \{1,2,3\}$ to aggregate hidden layers of the cross-attention layer to a single vector. The model can learn the relative importance of hidden states ($head^{mj}_i$) by learning the $\alpha_i$.
We compute $\alpha_i$ as: 
$\alpha_i = \mbox{softmax}(v^T \mbox{tanh}(W_h head^{mj}_i + b_h))$ 
where $W_h$ is the weight matrix, and $v$ and $b_h$ are the parameters of the network.
    

\subsection{Pretraining Approach}

We leverage the Video-Text Matching (VTM) pretraining task and also adopt analogous approaches for Video-Audio Matching (VAM) and Audio-Text Matching (ATM) to enhance the model's performance. Alongside this, we integrate contrastive learning during pretraining to enhance multimodal representation learning. Combining VTM, VAM, and ATM tasks with contrastive learning establishes a robust pretraining framework, aligning various modalities in a unified representation space. This joint pretraining approach enables the model to capture intricate relationships between video, audio, and text which leads to improved performance on downstream tasks. Following pretraining, we fine-tune the pretrained model for comic mischief detection and subtype classification. More details about pretraining approaches have been provided in the Appendix (\ref{pretraining-approaches}).


\subsection{Fine-Tuning}

In this section, we describe the process of fine-tuning the pretrained model for comic mischief prediction and classification. We present two task formulations. First, the binary classification task, aimed at identifying whether a clip contains comic mischief or not. Second, the multi-task classification for concurrently predicting all four subcategories.
We consider one MLP block (three FC layers with ReLU activation function and batch normalization) for binary classification and four MLP blocks for multi-task classification. All video, audio, and text representations are concatenated and passed through the MLP block to predict the output.
\subsubsection{Multi-task Learning}
It is well known that multi-task learning can improve performance when there is shared information between tasks, as seen in similar studies \cite{zhang2021none, zhang2023positive}. 
Although in our problem we only have a single task, 
the considered categories may be correlated, 
as all tasks care about the comic aspect accompanied by a kind of objectionable content. Therefore, we adopted a multi-task learning setting for training the model in the multi-label classification scenario. We added separate MLP blocks for each class as can be seen in Figure~\ref{Overal-structure}, part 4. 
In multi-task learning, each task will have its unique loss function, denoted by $L_i$. 
We weight each loss function and minimize a linear combination of these weighted losses; weights are learnable parameters that will be tuned during training $L_{total} = \sum_iw_iL_i $.

%

\section{Experimental evaluation}
\label{sec:experiments}
\subsection{Experimental Setup}
We utilize a subset (50K) of the HowTo100M \cite{miech2019howto100m} dataset for matching (VTM, VAM, and ATM) pretraining tasks and Kinetics-400~\cite{kay2017kinetics} for contrastive learning purposes in compliance with their policies. 

\noindent\textbf{Data Partitions: } We divided the dataset into three parts: 65\% for training, 10\% for validation, and 25\% for testing. 
Table \ref{comic-distr} shows the data distribution of the
dataset for each partition. 

\begin{table}[h!]
\centering
\scalebox{0.85}{
\centering
\begin{tabular}{|l|c|c|c|c|c|c|}
\hline
      & \multicolumn{1}{l|}{\textbf{MH}} & \multicolumn{1}{l|}{\textbf{SH}} & \multicolumn{1}{l|}{\textbf{GH}} & \multicolumn{1}{c|}{\textbf{S}} & \multicolumn{1}{l|}{\textbf{None}} & \multicolumn{1}{l|}{\textbf{\# Clips}} \\ \hline
Train        & 784 & 430 & 152 & 578 & 1296 & 2890   \\ \hline
Validation   & 103 & 79  & 26  & 74  &  199 & 432    \\ \hline
Test         & 231 & 99 & 36  & 168 & 725  & 1156    \\ \hline
\end{tabular}
}
\caption{Dataset partitions. Columns 2-4 indicate the number of samples per category; column 6 shows the total number of clips per partition.  MH, SH, GH, and S stand for mature humor, slapstick humor, gory humor, and sarcasm, respectively. Note that one clip can belong to more than one category. 
}
\label{comic-distr}
\end{table}

\noindent\textbf{Metrics: }
We report the F1 score with respect to the positive class for the binary model and the macro F1 score for the multi-task model.

\begin{table*}
\footnotesize
    \begin{minipage}[t]{0.31\linewidth}\centering
    \scalebox{0.7}{
    \begin{tabular}{|l|c|c|c|}
    \hline
    Models                          & Modality  & \begin{tabular}{@{}c@{}}BM  \\ F1-Score\end{tabular}  &  \begin{tabular}{@{}c@{}}MTM \\ Macro F1\end{tabular} \\ \hline
    \multirow{3}{*}{\begin{tabular}{@{}c@{}}Single \\ Modal\end{tabular}}   & T(+C)       & 60.60   & 60.15      \\ \cline{2-4} 
                                                                            & A           & 60.08   & 59.9       \\ \cline{2-4} 
                                                                            & V           & 59.93   & 58.27      \\ \hline
    \multirow{3}{*}{\begin{tabular}{@{}c@{}}Bi \\ Modal\end{tabular}}       & T(+C)+V     & 62.79   & 62.35      \\ \cline{2-4} 
                                                                            & A+V         & 63.66   & 63.01      \\ \cline{2-4} 
                                                                            & T(+C)+A     & 63.12   & 62.69      \\ \hline
    HICCAP                                                                 & \begin{tabular}{@{}c@{}}T(+C)+ \\ A+V\end{tabular}   & \textbf{71.72}   & \textbf{69.95}      \\ \hline
    \end{tabular}
    }
    \vspace{-0.1cm}
    a) Multimodality
    \vspace{-0.1cm}
    \label{ablation-multi-mod-bin}
    \end{minipage}
    \hfill
    \begin{minipage}[t]{0.33\linewidth}\centering
    \scalebox{0.75}{
    \begin{tabular}{|l|c|c|c|}
    \hline
    Models                          & Modality  & \begin{tabular}{@{}c@{}}BM \\ F1-Score \end{tabular}  &  \begin{tabular}{@{}c@{}}MTM \\ Macro F1\end{tabular} \\ \hline
    \begin{tabular}{@{}c@{}}Early \\ Fusion\end{tabular}        & \begin{tabular}{@{}c@{}}T(+C)+ \\ A+V\end{tabular}   & 63.93  & 62.12    \\ \hline
    \begin{tabular}{@{}c@{}}Late  \\ Fusion\end{tabular}        & \begin{tabular}{@{}c@{}}T(+C)+ \\ A+V\end{tabular}   & 64.69  & 62.86  \\ \hline
    HICCAP                                                      & \begin{tabular}{@{}c@{}}T(+C)+ \\ A+V\end{tabular}   & \textbf{71.72} & \textbf{69.95}    \\ \hline
    \end{tabular}
    }
    b) Hierarchical Cross Attention
    \label{ablation-HCA-bin}
    \vspace{-0.1cm}
    \end{minipage}
    \hfill
    \begin{minipage}[t]{0.33\linewidth}\centering
    \scalebox{0.75}{
    \begin{tabular}{|l|c|c|c|}
    \hline
    Models                          & Modality  & \begin{tabular}{@{}c@{}}BM \\ F1-Score \end{tabular}  &  \begin{tabular}{@{}c@{}}MTM \\ Macro F1\end{tabular} \\ \hline
    
    \multirow{2}{*}{\begin{tabular}{@{}c@{}}Single \\ Modal\end{tabular}}& T         & 59.12   & 58.85    \\ \cline{2-4} 
                                                                         & T(+C)     & 60.60   & 60.15     \\ \hline
    HCA                                                                  & T+A+V     & 70.36   & 69.56   \\ \hline
    HICCAP                      & \begin{tabular}{@{}c@{}}T(+C)+ \\ A+V\end{tabular} & \textbf{71.72}   & \textbf{69.95}    \\ \hline
    \end{tabular}
    }
    c) Adapting caption generation
    \label{ablation-cap-bin}
    \vspace{-0.1cm}
    \end{minipage}
\caption{Ablation study on the effect of a) Multimodality, b) Hierarchical Cross Attention, and c) Adapting caption generation, for comic mischief detection in the binary model (BM) (Column 3) and type prediction in the multi-task model (MTM)(Column 4). ‘T’, ‘A’, ‘V’, and 'C' stand for text, audio, video, and caption.}
\label{ablation-multi-HCA-cap-bin}
\end{table*}

\subsection{Ablation and Analysis}
\label{ablation-study}
In this section, we carry out a comprehensive set of experiments to assess the performance of our proposed model in comparison to a range of baseline models, focusing on two tasks: first, the binary classification of comic mischief presence, and second, the multi-task prediction of specific comic mischief subcategories.
In our evaluation, we have selected several baseline systems for comparison. The baseline categories include: 1) unimodal systems 
, 2) bi-modal approaches, and 3) models that consider all three modalities but do not employ a hierarchical cross-attention mechanism.
The rationale behind selecting these baselines is to demonstrate the importance of both multi-modal integration and the hierarchical cross-attention mechanism for accurately predicting comic mischief content and its type. 
Additionally, we conducted experiments to highlight the significance of pretraining techniques.

\textbf{Impact of Multimodality:}
We provide a comparative analysis between the single-modal model, the Bi-modal model, and our HICCAP for binary and multi-task prediction. In single-modal models, we employ a pretrained BERT with attention and fully connected (FC) layers for text, and deploy LSTM integrated with attention and FC layers for both audio and video modalities. The Bi-modal models integrate the original cross-attention mechanism limited to two modalities. Our results in Table \ref{ablation-multi-HCA-cap-bin}.a indicate that HICCAP significantly improves upon the best-performing monomodal and Bi-modal cross-attention models by 11.12\% and 8.06\%, respectively, in terms of F1 score, for the binary classification task. Also in multi-task experiments, HICCAP outperforms these models by 9.8\% and 6.94\%, respectively, based on the macro F1. The improvements are statistically significant with a p-value < 0.001 according to the Mcnemar significance test.


\textbf{Impact of Hierarchical Cross-Attention:} 
To assess the contribution of hierarchical cross-attention, we compare HICCAP against multimodal concatenation models, including early and late fusion techniques. The early fusion strategy integrates modalities by concatenating feature vectors, while the late fusion trains each modality independently and combines them at the decision level by averaging class probabilities. 
Table \ref{ablation-multi-HCA-cap-bin}.b showcases that HICCAP outperforms both early and late fusion models, by 7.79\% and 7.03\%, respectively, based on the F1 score for binary classification.  Besides, in this table for multi-task prediction, HICCAP improves the performance of early and late fusion models by 7.83\% and 7.09\%, respectively, based on macro F1 scores.


\textbf{Impact of Adapting Caption Generation:} 
To illustrate the influence of captioning on videos without subtitles, we incorporated captions into both the text single modality and hierarchical cross-attention (HCA) models for binary and multi-task prediction. The results in Table \ref{ablation-multi-HCA-cap-bin}.c reveal improvements for both models, with HICCAP surpassing the HCA model by 1.36\% in terms of F1 score, for binary classification. Also, the results show applying the caption generation technique improves the performance of the HCA model by 0.39\% for multi-task prediction. However, adding captions is not as effective as in the binary model.

%


\begin{table*}
\footnotesize
    \begin{minipage}[t]{0.32\linewidth}\centering
    \scalebox{0.73}{
    \begin{tabular}{|l|c|c|c|}
    \hline
    Models                          & Modality  & \begin{tabular}{@{}c@{}}BM \\ F1-Score\end{tabular}  &  \begin{tabular}{@{}c@{}}MTM \\ Macro F1\end{tabular} \\ \hline
    
    \begin{tabular}{@{}l@{}}w/o \\ Pretrain\end{tabular}              & \begin{tabular}{@{}c@{}}T(+C)+ \\ A+V\end{tabular}  & 71.72  & 69.95  \\ \hline
    Matching                                                             & \begin{tabular}{@{}c@{}}T(+C)+ \\ A+V\end{tabular}  & 72.24  & 72.48  \\ \hline
    CL                                                                & \begin{tabular}{@{}c@{}}T(+C)+ \\ A+V\end{tabular}  & 73.12  & 73.30   \\ \hline
    \begin{tabular}{@{}l@{}}Hybrid CL \\ +Matching\end{tabular}   & \begin{tabular}{@{}c@{}}T(+C)+ \\ A+V\end{tabular}  & \textbf{74.96}  & \textbf{74.13}   \\ \hline
    \end{tabular}
    }
    \caption{Ablation study on the effect of pretraining techniques for comic mischief detection and type classification.}
    \label{ablation-pre-bin}
    \end{minipage}
    \hfill
    \begin{minipage}[t]{0.4\linewidth}\centering
    \scalebox{0.7}{
    \begin{tabular}{|l|c|c|c|c|c|c|}
    \hline
    Method       &  \begin{tabular}{@{}c@{}}F1 \\ MH\end{tabular}   & \begin{tabular}{@{}c@{}}F1 \\ GH\end{tabular}     & \begin{tabular}{@{}c@{}}F1 \\ SH\end{tabular}    & \begin{tabular}{@{}c@{}}F1 \\ S\end{tabular} & \begin{tabular}{@{}c@{}} Macro \\ F1\end{tabular} \\ \hline
    \begin{tabular}[c]{@{}l@{}}Late fusion \\ per task\end{tabular}    & 67.05  & 42.08 & 51.83            & 69.83            & 57.69                            \\ \hline
    \begin{tabular}[c]{@{}l@{}}GMU fusion \\ per task\end{tabular}       & 49.14  & 46.73 & 45.93           & 38.37           & 45.04                  \\ \hline
    \begin{tabular}{@{}l@{}}HICCAP \\ per task\end{tabular}             & 79.44  & 73.59 & 43.80            & 74.42            & 67.81                           \\ \hline
    \begin{tabular}{@{}l@{}}Multi-label \\ HICCAP\end{tabular}       & 78.64  & 57.54 & 77.36              & 58.83            & 68.09                            \\ \hline
    \begin{tabular}{@{}l@{}}Multi-task \\ HICCAP\end{tabular}        & 76.83  & 65.73 & 62.34              & 74.93            & \textbf{69.95}    \\ \hline
    \end{tabular}%
    }
    \vspace{-0.1cm}
    \caption{Comparing a) multi-task model with multi-label and single-task models; b) HCA per task with late and GMU fusion.
    \label{comic-result3}}
    \end{minipage}
    \hfill
    \begin{minipage}[t]{0.24\linewidth}\centering
    \scalebox{0.73}{
    \begin{tabular}{|l|c|c|}
    \hline
    Models & Modality  &   F1-score  \\ \hline
    
    GMU                      & \begin{tabular}{@{}c@{}}T(+C)+ \\ A+V\end{tabular}    & 46.90  \\ \hline
    \multirow{2}{*}{LXMERT}  & T(+C)+V      & 64.24   \\ \cline{2-3} 
                             & T(+C)+A      & 63.61    \\ \hline
    X-Clip                   & T(+C)+V      & 68.80     \\ \hline
    HICCAP                   & \begin{tabular}{@{}c@{}}T(+C)+ \\ A+V\end{tabular}    & 71.72     \\ \hline
    \begin{tabular}{@{}l@{}}Pretrain \\ HICCAP\end{tabular}  & \begin{tabular}{@{}c@{}}T(+C)+ \\ A+V\end{tabular}  & \textbf{74.96}  \\ \hline
    \end{tabular}
    }
    \vspace{-0.1cm}
    \caption{Comparison of HICCAP with previous methods for binary comic mischief detection.}
    \label{comic-result-SOTA}
    \end{minipage}
\end{table*}

\textbf{Impact of Pretraining Techniques:} 
To show the importance of pretraining, we conduct experiments focusing on matching pretraining only, contrastive learning (CL) only, and a combination of both. The data in Table \ref{ablation-pre-bin} highlights that HICCAP with hybrid pretraining surpasses both the non-pretrained variant and those with matching and CL pretraining, showing gains of 3.24\%, 2.72\%, and 1.84\%, respectively, based on the F1 score, for binary classification. Furthermore, the findings in this table demonstrate that for multi-task prediction employing a hybrid pretraining method leads to a significant improvement in HICCAP performance, in terms of macro F1 score.

%

\textbf{Impact of Multi-Task Learning:} 
Table~\ref{comic-result3} compares the result of the multi-task setting of our model with the four single-task hierarchical cross-attention models. All the models are exactly the same, but we trained them for each task separately. The multi-task model works better overall based on the macro F1. Moreover, to show a hierarchical cross-attention mechanism in the single-task models is a reasonable approach, we compare the single-task models with the late fusion and intermediate GMU fusion models. The hierarchical cross-attention model outperforms late fusion and GMU fusion by 10.12\% and 22.77\%, respectively, based on macro F1. 
Also, as shown in this table, the multi-task approach performs better than the multi-label approach by 1.86\% in terms of macro F1. 
Given the superior performance of the multi-task strategy, we decided to adopt this approach for the architecture of our proposed HICCAP model. This choice was motivated by our aim to maximize the effectiveness and efficiency of the model in addressing the complexities of the underlying tasks.


\textbf{Model Evaluation:}
We consider the following multimodal methods to evaluate the binary model:

    \noindent\textbf{LXMERT \cite{tan2019lxmert}:} 
To make this framework compatible with our work, we pass the video-encoded vector (I3D features) to the model rather than object vectors.

    \noindent\textbf{X-CLIP\cite{ni2022expanding}:} 
In order to incorporate X-CLIP with our structure, we utilized the checkpoint provided by the authors for zero-shot training. Subsequently, we fine-tuned and assessed the model using the comic mischief dataset.

    \noindent\textbf{Intermediate fusion with GMU:} As with the MM-trailer model \cite{shafaei2021case}, 
we train the model for each modality separately, then save the final layer before the classification layer, and ultimately use the GMU model to integrate the output from all modalities.
Table~\ref{comic-result-SOTA} shows the performance of the binary model for the comic mischief detection task. The table shows that the best results were obtained from the HICCAP with the hybrid pertaining technique. To fairly evaluate the HICCAP mechanism and compare it to other methods, we also report results for models without pretraining. 
The results show that HICCAP improves the best intermediate GMU fusion~\cite{shafaei2021case}, LXMERT~\cite{tan2019lxmert}, and X-CLIP~\cite{ni2022expanding} models by 24.82\%, 7.48\%, and 2.92\%, respectively, based on F1 score. 

\vspace{-0.2cm}
\subsection{Comparison with State-of-the-Art}
\vspace{-0.1cm}
We assess the performance of our HICCAP on two multimodal datasets designed for activity recognition: UCF-101 \cite{soomro2012ucf101} and HMDB51 \cite{kuehne2011hmdb}, as well as on the XD-Violence \cite{wu2020not} dataset tailored for anomaly detection.
For caption generation on these datasets, we employ the prompt generation technique presented in X-Clip \cite{ni2022expanding}. 
In this method, the authors enhanced the original text encoder, initially trained for language-image tasks, with a video-specific prompting mechanism. By utilizing video content features, they aimed to optimize the text prompting process, emphasizing that proper contextual information boosts recognition. 
\vspace{-0.2cm}
\subsubsection{Result on UCF101 and HMDB51 Datasets}
\vspace{-0.1cm}
Table \ref{ucf101-hmdb51-result} summarizes the performance comparison between the proposed method and other state-of-the-art methods on the UCF101 and HMDB51 datasets. In the presented table, our method outperforms the majority of previously established techniques in terms of top-1 accuracy. Notably, there is a single method, VideoMAE V2~\cite{wang2023videomae} that our proposal does not outperform. 
However, it's crucial to highlight that even in this case, our performance is comparable to this paper. Importantly, our method achieves this with a more straightforward strategy and a model with significantly fewer parameters. This underscores the efficacy and efficiency of our approach, demonstrating that competitive outcomes can be realized without the need for complex systems or excessive model parameters.



\begin{table}[h!]
\centering
\scalebox{0.68}{
\begin{tabular}{|l|c|c|c|}
    \hline
    Method & Modality & UCF101 & HMDB51 \\
    \hline
    VideoMoCo \cite{pan2021videomoco}& V     & 78.7 & 49.2 \\\hline
    Vi2CLR \cite{diba2021vi2clr}& V          & 89.1 & 55.7 \\\hline
    CVRL \cite{qian2021spatiotemporal}& V    & 94.4 & 70.6 \\\hline
    CORPf \cite{hu2021contrast}& V           & 93.5 & 68.0 \\\hline
    MIL-NCE \cite{miech2020end}& T+V         & 91.3 & 61.0 \\\hline
    MMV \cite{alayrac2020self}& T+A+V        & 92.5 & 69.6 \\\hline
    CPD \cite{li2020learning}& T+V           & 92.8 & 63.8 \\\hline
    ELO \cite{piergiovanni2020evolving}& A+V & 93.8 & 67.4 \\\hline
    XDC \cite{alwassel2020self}& A+V         & 94.2 & 67.1 \\\hline
    GDT \cite{patrick2020multi}& A+V         & 95.2 & 72.8 \\\hline
    VideoMAE V1\cite{tong2022videomae}& V    & 96.1 & 73.3 \\\hline
    VideoMAE V2\cite{wang2023videomae}& V    & 99.6 & 88.1 \\\hline
    Ours & T+A+V                             & \textbf{98.87}& \textbf{76.64}   \\\hline
\end{tabular}
}
\caption{Comparison with the UCF101 and HMDB51 datasets.‘T’, ‘A’, ‘V’, and 'C' stand for text, audio, video, and caption.} 
\label{ucf101-hmdb51-result}
\end{table}

\subsubsection{Result on XD-Violence Dataset}
In this section, we applied our method to the XD-Violence dataset. Due to the lack of access to part of the provided vector features, we regenerate the I3D and VGGish features using the provided training and test videos. Similar to \cite{wu2021learning}, we use the frame-level precision-recall curve (PRC) and corresponding area under the curve (average precision, AP) and report the results in Table \ref{XD-result}. As shown in this table, our method achieves the new state-of-the-art performance of 92.17\% AP and gains clear improvements when compared with previous SOTA methods in terms of AP.

\begin{table}[h!]
\centering
\scalebox{0.85}{
\begin{tabular}{|l|c|c|}
    \hline
    Method & Modality & AP (\%) \\
    \hline
    AVVD \cite{wu2022weakly}                  & V     & 78.10  \\\hline
    NG-MIL \cite{park2023normality}           & V     & 78.51  \\\hline
    MGFN \cite{chen2023mgfn}                  & V     & 80.11  \\\hline
    SR3 \cite{wu2022self}                     & V     & 80.26  \\\hline
    CU-Net \cite{zhang2023exploiting}         & A+V   & 81.43  \\\hline
    DMU \cite{zhou2023dual}                   & A+V   & 81.77   \\\hline
    CLIP-TSA \cite{joo2023clip}               & V     & 82.19  \\\hline
    MACIL-SD \cite{yu2022modality}            & A+V   & 83.4  \\\hline
    DDl \cite{pu2022audio}                    & A+V   & 83.54  \\\hline
    Song et al. \cite{fan2023weakly}          & A+V   & 84.23  \\\hline
    VadCLIP \cite{wu2023vadclip}              & V+T   & 84.51  \\\hline
    Ours                                      & T+A+V & \textbf{92.17}  \\\hline
\end{tabular}
}
\caption{Comparison with SOTA on the XD-Violence dataset. ‘T’, ‘A’, ‘V’, and 'C' stand for text, audio, video, and caption.}
\label{XD-result}
\end{table}

\subsection{Error Analysis}
In the analysis of misclassified instances of comic mischief, approximately 80\% belong to a single category (such as sarcasm exclusively), while the remainder involved multiple categories. This observation aligns with the anticipation that videos with multiple categories tend to be more accurately identified as instances of comic mischief. Also, within the "none" category, about 35\% of predictions were incorrectly predicted. This could be due to the presence of mature or violent content lacking humorous elements, or possibly from the smaller number of "none" examples, complicating the model's ability to discern a clear pattern.

Our investigation into the reliance of the model on various modalities revealed enlightening insights, particularly with respect to sarcasm—a category predominantly inferred from textual dialogues. An experimental manipulation involved masking the video and audio components for clips categorized under sarcasm, leading to mispredictions by our model. This outcome underscores a critical observation: despite sarcasm's heavy reliance on textual cues within dialogues, the integration of video and audio modalities plays a non-negligible role in enhancing the model's predictive performance. 
Extending this line of inquiry to the gory and slapstick humor categories, which are primarily video-centric, we conducted similar experiments by masking text and audio modalities to observe the impact on multimodal prediction capabilities. The modifications led to a significant degradation in the model's performance for these categories. 
These mispredictions highlight the indispensable value of multimodal data fusion in understanding complex comic mischief categories, thereby emphasizing the need for a comprehensive approach in the processing and interpretation of multimodal information for accurate content categorization.




\section{Related Work}
\label{sec-back}
\vspace{-0.1cm}
\textbf{Multimodal Machine Learning (MML).} MML \cite{baltruvsaitis2018multimodal,xu2022multimodal} has gained considerable attention in recent decades as a vital research area. 
In recent literature, numerous models have been proposed to analyze multimodal information in videos, taking into account the complex interplay between visual, auditory, and textual data ~\cite{seo2022end, man2022scenario, basak2022union, sun2022human}. 
In~\cite{akbari2021vatt}, the authors present a Video-Audio-Text Transformer (VATT) for learning multimodal representations from unlabeled data using convolution-free transformer architectures.
VATLM ~\cite{zhu2022vatlm} combines video, audio, and text modalities by utilizing a unified transformer tokenizer following modality-specific encoders and subsequently performing masked prediction on the integrated tokens. 
Compared with previous work, HICCAP introduces a novel method that combines various Hierarchical multimodal Cross-Attention (HCA) mechanisms. Each HCA learns cross-attention between one modality and the other two, producing unique representations for each modality, ultimately leading to improved model performance.

\textbf{Cross-Modal Pretraining Tasks.}
Several studies have focused on pretraining tasks to warm-start multimodal model parameters, aiming to boost performance in downstream tasks. One popular pretraining task is Image-Text Matching (ITM) which aims at grasping the coarse-grained correlation between images and texts~\cite{huang2021seeing,li2020unicoder,du2022survey}. Recent works like~\cite{li2021align} have employed a multimodal encoder with cross-attention for the ITM task. Contrastive learning (CL) has also seen significant advancements in representation learning, both in unimodal~\cite{chen2020simple, gao2021simcse, oord2018representation} and multimodal contexts~\cite{li2022blip, li2023blip, yuan2021multimodal, ramesh2021zero, udandarao2020cobra, goel2022cyclip}. Notable implementations, such as CLIP~\cite{radford2021learning}, align text-image pairs, while its extension, X-Clip~\cite{ni2022expanding}, focuses on video recognition. LAVA~\cite{gurram2022lava} delves into video-audio-text multimodal pretraining. 
Recent advancements in multimodal learning have highlighted the benefits of hybrid pretraining techniques. Notably, BLIP\cite{li2022blip} introduces a combination of image-text matching and image-text contrastive pretraining. Further, BLIP-2 \cite{li2023blip} expands the approach by jointly optimizing three objectives: Image-Text Matching, Image-Text Contrastive Learning, and Image-Grounded Text Generation.
In this paper, 
HICCAP employs a multimodal video-audio-text CL strategy, combined with video-audio-text matching pretraining techniques, aiming for the integrated learning of language, audio, and video representations.

\section{Conclusion}
\label{sec:conclusion}
We introduced the task of labeling comic mischief in videos and released a new dataset for this purpose. We also proposed a novel model named ''HIerarchical Cross attention model with CAPtions'' (HICCAP). HICCAP uses hierarchical cross-attention modules to capture interactions between the three modalities. We pretrained the HICCAP model with hybrid modality matching and contrastive learning before fine-tuning it for binary and multi-task classifications. The experiments indicate that our method outperforms both the baseline and other reference models.
Our main conclusions are the following:
\begin{compactitem}
    \item \emph{Detecting comic mischief in videos is a feasible task for multimodal learning.} 
    Likewise, there is still too much room for improvement in this relevant task, therefore, we foresee it will be of interest to the community. 
    \item \emph{Hierarchical cross-attention (HCA) obtained better performance than standard cross-attention. } Contrary to most work on multimodal learning, HCA is able to capture dependencies across all of the considered modalities, this way of modeling dependencies resulted in better performance when compared to models that only implemented self and bi-modal cross-attention.  
    \item \emph{Both,  multimodal matching pretraining and contrastive learning pretraining proved to be useful to warm-start the model, and their combination further boosted the model's performance.} We show that the adoption of the hybrid pretraining process resulted in the best performance for our HICCAP model. 
\end{compactitem}

Future work includes the exploration of hierarchical cross-attention in the context of other multimodal learning models. Also, we would like to develop explainable models for detecting comic mischief content. 

\section{Acknowledgments}
\label{sec:acknowledgments}
We're grateful to the anonymous LREC-COLING reviewers for their valuable comments. We appreciate the support from Ioannis Kakadiaris and Christos Smailis during the dataset annotation process. We also thank Jared Suchomel, Alli Brophy, Lyndon Lam, Connor Maguire, and Kelly Couvrette for their assistance with dataset annotation. This research was partially funded by the National Science Foundation, under awards 1910192 and 2106892.



\nocite{*}
\section{Bibliographical References}\label{sec:reference}

\bibliographystyle{lrec-coling2024-natbib}
\bibliography{lrec-coling2024-example}

\begin{thebibliography}{93}
\expandafter\ifx\csname natexlab\endcsname\relax\def\natexlab#1{#1}\fi

\bibitem[{Abu-El-Haija et~al.(2016)Abu-El-Haija, Kothari, Lee, Natsev,
  Toderici, Varadarajan, and Vijayanarasimhan}]{abu2016youtube}
Sami Abu-El-Haija, Nisarg Kothari, Joonseok Lee, Paul Natsev, George Toderici,
  Balakrishnan Varadarajan, and Sudheendra Vijayanarasimhan. 2016.
\newblock Youtube-8m: A large-scale video classification benchmark.
\newblock \emph{arXiv preprint arXiv:1609.08675}.

\bibitem[{Akbari et~al.(2021)Akbari, Yuan, Qian, Chuang, Chang, Cui, and
  Gong}]{akbari2021vatt}
Hassan Akbari, Liangzhe Yuan, Rui Qian, Wei-Hong Chuang, Shih-Fu Chang, Yin
  Cui, and Boqing Gong. 2021.
\newblock Vatt: Transformers for multimodal self-supervised learning from raw
  video, audio and text.
\newblock \emph{Advances in Neural Information Processing Systems},
  34:24206--24221.

\bibitem[{Alayrac et~al.(2020)Alayrac, Recasens, Schneider, Arandjelovi{\'c},
  Ramapuram, De~Fauw, Smaira, Dieleman, and Zisserman}]{alayrac2020self}
Jean-Baptiste Alayrac, Adria Recasens, Rosalia Schneider, Relja
  Arandjelovi{\'c}, Jason Ramapuram, Jeffrey De~Fauw, Lucas Smaira, Sander
  Dieleman, and Andrew Zisserman. 2020.
\newblock Self-supervised multimodal versatile networks.
\newblock \emph{Advances in Neural Information Processing Systems}, 33:25--37.

\bibitem[{Alwassel et~al.(2020)Alwassel, Mahajan, Korbar, Torresani, Ghanem,
  and Tran}]{alwassel2020self}
Humam Alwassel, Dhruv Mahajan, Bruno Korbar, Lorenzo Torresani, Bernard Ghanem,
  and Du~Tran. 2020.
\newblock Self-supervised learning by cross-modal audio-video clustering.
\newblock \emph{Advances in Neural Information Processing Systems},
  33:9758--9770.

\bibitem[{Anderson et~al.(2003)Anderson, Berkowitz, Donnerstein, Huesmann,
  Johnson, Linz, Malamuth, and Wartella}]{anderson2003influence}
Craig~A Anderson, Leonard Berkowitz, Edward Donnerstein, L~Rowell Huesmann,
  James~D Johnson, Daniel Linz, Neil~M Malamuth, and Ellen Wartella. 2003.
\newblock The influence of media violence on youth.
\newblock \emph{Psychological science in the public interest}, 4(3):81--110.

\bibitem[{Bahdanau et~al.(2014)Bahdanau, Cho et~al.}]{bahdanau2014neural}
Dzmitry Bahdanau, Kyunghyun Cho, et~al. 2014.
\newblock Neural machine translation by jointly learning to align and
  translate.
\newblock \emph{arXiv preprint arXiv: 1409.0473}.

\bibitem[{Baltru{\v{s}}aitis et~al.(2018)Baltru{\v{s}}aitis, Ahuja, and
  Morency}]{baltruvsaitis2018multimodal}
Tadas Baltru{\v{s}}aitis, Chaitanya Ahuja, and Louis-Philippe Morency. 2018.
\newblock Multimodal machine learning: A survey and taxonomy.
\newblock \emph{IEEE transactions on pattern analysis and machine
  intelligence}, 41(2):423--443.

\bibitem[{Basak et~al.(2022)Basak, Kundu, Singh, Ijaz, Wo{\'z}niak, and
  Sarkar}]{basak2022union}
Hritam Basak, Rohit Kundu, Pawan~Kumar Singh, Muhammad~Fazal Ijaz, Marcin
  Wo{\'z}niak, and Ram Sarkar. 2022.
\newblock A union of deep learning and swarm-based optimization for 3d human
  action recognition.
\newblock \emph{Scientific Reports}, 12(1):5494.

\bibitem[{Benaim et~al.(2020)Benaim, Ephrat, Lang, Mosseri, Freeman,
  Rubinstein, Irani, and Dekel}]{benaim2020speednet}
Sagie Benaim, Ariel Ephrat, Oran Lang, Inbar Mosseri, William~T Freeman,
  Michael Rubinstein, Michal Irani, and Tali Dekel. 2020.
\newblock Speednet: Learning the speediness in videos.
\newblock In \emph{Proceedings of the IEEE/CVF Conference on Computer Vision
  and Pattern Recognition}, pages 9922--9931.

\bibitem[{Blackford et~al.(2011)Blackford, Gentry, Harrison, and
  Carlson}]{blackford2011prevalence}
Benjamin~J Blackford, James Gentry, Robert~L Harrison, and Les Carlson. 2011.
\newblock The prevalence and influence of the combination of humor and violence
  in super bowl commercials.
\newblock \emph{Journal of Advertising}, 40(4):123--134.

\bibitem[{Carreira and Zisserman(2017)}]{carreira2017quo}
Joao Carreira and Andrew Zisserman. 2017.
\newblock Quo vadis, action recognition? a new model and the kinetics dataset.
\newblock In \emph{proceedings of the IEEE Conference on Computer Vision and
  Pattern Recognition}, pages 6299--6308.

\bibitem[{Chang and Bushman(2019)}]{chang2019effect}
Justin~H Chang and Brad~J Bushman. 2019.
\newblock Effect of exposure to gun violence in video games on children’s
  dangerous behavior with real guns: a randomized clinical trial.
\newblock \emph{JAMA network open}, 2(5):e194319--e194319.

\bibitem[{Chang et~al.(2021)Chang, Li, Shen, Feng, and
  Zhou}]{chang2021contrastive}
Shuning Chang, Yanchao Li, Shengmei Shen, Jiashi Feng, and Zhiying Zhou. 2021.
\newblock Contrastive attention for video anomaly detection.
\newblock \emph{IEEE Transactions on Multimedia}, 24:4067--4076.

\bibitem[{Chen et~al.(2021{\natexlab{a}})Chen, Rouditchenko, Duarte, Kuehne,
  Thomas, Boggust, Panda, Kingsbury, Feris, Harwath
  et~al.}]{chen2021multimodal}
Brian Chen, Andrew Rouditchenko, Kevin Duarte, Hilde Kuehne, Samuel Thomas,
  Angie Boggust, Rameswar Panda, Brian Kingsbury, Rogerio Feris, David Harwath,
  et~al. 2021{\natexlab{a}}.
\newblock Multimodal clustering networks for self-supervised learning from
  unlabeled videos.
\newblock In \emph{Proceedings of the IEEE/CVF International Conference on
  Computer Vision}, pages 8012--8021.

\bibitem[{Chen et~al.(2021{\natexlab{b}})Chen, Huang, He, Long, Zeng, Wen, Tan,
  and Gan}]{chen2021rspnet}
Peihao Chen, Deng Huang, Dongliang He, Xiang Long, Runhao Zeng, Shilei Wen,
  Mingkui Tan, and Chuang Gan. 2021{\natexlab{b}}.
\newblock Rspnet: Relative speed perception for unsupervised video
  representation learning.
\newblock In \emph{Proceedings of the AAAI Conference on Artificial
  Intelligence}, volume~35, pages 1045--1053.

\bibitem[{Chen et~al.(2020)Chen, Kornblith, Norouzi, and
  Hinton}]{chen2020simple}
Ting Chen, Simon Kornblith, Mohammad Norouzi, and Geoffrey Hinton. 2020.
\newblock A simple framework for contrastive learning of visual
  representations.
\newblock In \emph{International conference on machine learning}, pages
  1597--1607. PMLR.

\bibitem[{Chen et~al.(2022)Chen, Kang, Wang, Li, and Lu}]{chen2022efficient}
Xin Chen, Ben Kang, Dong Wang, Dongdong Li, and Huchuan Lu. 2022.
\newblock Efficient visual tracking via hierarchical cross-attention
  transformer.
\newblock In \emph{European Conference on Computer Vision}, pages 461--477.
  Springer.

\bibitem[{Chen et~al.(2023)Chen, Liu, Zhang, Fok, Qi, and Wu}]{chen2023mgfn}
Yingxian Chen, Zhengzhe Liu, Baoheng Zhang, Wilton Fok, Xiaojuan Qi, and
  Yik-Chung Wu. 2023.
\newblock Mgfn: Magnitude-contrastive glance-and-focus network for
  weakly-supervised video anomaly detection.
\newblock In \emph{Proceedings of the AAAI Conference on Artificial
  Intelligence}, volume~37, pages 387--395.

\bibitem[{Devlin et~al.(2018)Devlin, Chang, Lee, and
  Toutanova}]{devlin2018bert}
J.~Devlin, M.W. Chang, K.~Lee, and K.~Toutanova. 2018.
\newblock Bert: Pre-training of deep bidirectional transformers for language
  understanding.
\newblock \emph{arXiv preprint arXiv:1810.04805}.

\bibitem[{Diba et~al.(2021)Diba, Sharma, Safdari, Lotfi, Sarfraz, Stiefelhagen,
  and Van~Gool}]{diba2021vi2clr}
Ali Diba, Vivek Sharma, Reza Safdari, Dariush Lotfi, Saquib Sarfraz, Rainer
  Stiefelhagen, and Luc Van~Gool. 2021.
\newblock Vi2clr: Video and image for visual contrastive learning of
  representation.
\newblock In \emph{Proceedings of the IEEE/CVF international conference on
  computer vision}, pages 1502--1512.

\bibitem[{Dietz and Strasburger(1991)}]{dietz1991children}
William~H Dietz and Victor~C Strasburger. 1991.
\newblock Children, adolescents, and television.
\newblock \emph{Current problems in pediatrics}, 21(1):8--31.

\bibitem[{Du et~al.(2022)Du, Liu, Li, and Zhao}]{du2022survey}
Yifan Du, Zikang Liu, Junyi Li, and Wayne~Xin Zhao. 2022.
\newblock A survey of vision-language pre-trained models.
\newblock \emph{arXiv preprint arXiv:2202.10936}.

\bibitem[{Fan et~al.(2023)Fan, Yu, Lu, and Han}]{fan2023weakly}
Yidan Fan, Yongxin Yu, Wenhuan Lu, and Yahong Han. 2023.
\newblock Weakly-supervised video anomaly detection with snippet anomalous
  attention.
\newblock \emph{arXiv preprint arXiv:2309.16309}.

\bibitem[{Feichtenhofer et~al.(2021)Feichtenhofer, Fan, Xiong, Girshick, and
  He}]{feichtenhofer2021large}
Christoph Feichtenhofer, Haoqi Fan, Bo~Xiong, Ross Girshick, and Kaiming He.
  2021.
\newblock A large-scale study on unsupervised spatiotemporal representation
  learning.
\newblock In \emph{Proceedings of the IEEE/CVF Conference on Computer Vision
  and Pattern Recognition}, pages 3299--3309.

\bibitem[{Gao et~al.(2021)Gao, Yao, and Chen}]{gao2021simcse}
Tianyu Gao, Xingcheng Yao, and Danqi Chen. 2021.
\newblock Simcse: Simple contrastive learning of sentence embeddings.
\newblock \emph{arXiv preprint arXiv:2104.08821}.

\bibitem[{Goel et~al.(2022)Goel, Bansal, Bhatia, Rossi, Vinay, and
  Grover}]{goel2022cyclip}
Shashank Goel, Hritik Bansal, Sumit Bhatia, Ryan Rossi, Vishwa Vinay, and
  Aditya Grover. 2022.
\newblock Cyclip: Cyclic contrastive language-image pretraining.
\newblock \emph{Advances in Neural Information Processing Systems},
  35:6704--6719.

\bibitem[{Gurram et~al.(2022)Gurram, Chan, Fang, and Canny}]{gurram2022lava}
Sumanth Gurram, David Chan, Andy Fang, and John Canny. 2022.
\newblock Lava: Language audio vision alignment for data-efficient video
  pre-training.
\newblock In \emph{First Workshop on Pre-training: Perspectives, Pitfalls, and
  Paths Forward at ICML 2022}. PMLR.

\bibitem[{Gutmann and Hyv{\"a}rinen(2010)}]{gutmann2010noise}
Michael Gutmann and Aapo Hyv{\"a}rinen. 2010.
\newblock Noise-contrastive estimation: A new estimation principle for
  unnormalized statistical models.
\newblock In \emph{Proceedings of the thirteenth international conference on
  artificial intelligence and statistics}, pages 297--304. JMLR Workshop and
  Conference Proceedings.

\bibitem[{Han et~al.(2020{\natexlab{a}})Han, Xie, and
  Zisserman}]{han2020memory}
Tengda Han, Weidi Xie, and Andrew Zisserman. 2020{\natexlab{a}}.
\newblock Memory-augmented dense predictive coding for video representation
  learning.
\newblock In \emph{European conference on computer vision}, pages 312--329.
  Springer.

\bibitem[{Han et~al.(2020{\natexlab{b}})Han, Xie, and Zisserman}]{han2020self}
Tengda Han, Weidi Xie, and Andrew Zisserman. 2020{\natexlab{b}}.
\newblock Self-supervised co-training for video representation learning.
\newblock \emph{Advances in Neural Information Processing Systems},
  33:5679--5690.

\bibitem[{Hanewinkel et~al.(2014)Hanewinkel, Sargent, Hunt, Sweeting, Engels,
  Scholte, Mathis, and Morgenstern}]{hanewinkel2014portrayal}
R.~Hanewinkel, J.~D. Sargent, K.~Hunt, H.~Sweeting, R.~C. Engels, R.~H.
  Scholte, E.~Mathis, F.and~Florek, and M.~Morgenstern. 2014.
\newblock Portrayal of alcohol consumption in movies and drinking initiation in
  low-risk adolescents.
\newblock \emph{Pediatrics}, 133(6):973--982.

\bibitem[{Hasan et~al.(2016)Hasan, Choi, Neumann, Roy-Chowdhury, and
  Davis}]{hasan2016learning}
Mahmudul Hasan, Jonghyun Choi, Jan Neumann, Amit~K Roy-Chowdhury, and Larry~S
  Davis. 2016.
\newblock Learning temporal regularity in video sequences.
\newblock In \emph{Proceedings of the IEEE conference on computer vision and
  pattern recognition}, pages 733--742.

\bibitem[{Hershey et~al.(2017)Hershey, Chaudhuri, Ellis, Gemmeke, Jansen,
  Moore, Plakal, Platt, Saurous, Seybold et~al.}]{hershey2017cnn}
Shawn Hershey, Sourish Chaudhuri, Daniel~PW Ellis, Jort~F Gemmeke, Aren Jansen,
  R~Channing Moore, Manoj Plakal, Devin Platt, Rif~A Saurous, Bryan Seybold,
  et~al. 2017.
\newblock Cnn architectures for large-scale audio classification.
\newblock In \emph{2017 ieee international conference on acoustics, speech and
  signal processing (icassp)}, pages 131--135. IEEE.

\bibitem[{Hu et~al.(2021)Hu, Shao, Liu, Raj, Savvides, and
  Shen}]{hu2021contrast}
Kai Hu, Jie Shao, Yuan Liu, Bhiksha Raj, Marios Savvides, and Zhiqiang Shen.
  2021.
\newblock Contrast and order representations for video self-supervised
  learning.
\newblock In \emph{Proceedings of the IEEE/CVF International Conference on
  Computer Vision}, pages 7939--7949.

\bibitem[{Huang et~al.(2021)Huang, Zeng, Huang, Liu, Fu, and
  Fu}]{huang2021seeing}
Zhicheng Huang, Zhaoyang Zeng, Yupan Huang, Bei Liu, Dongmei Fu, and Jianlong
  Fu. 2021.
\newblock Seeing out of the box: End-to-end pre-training for vision-language
  representation learning.
\newblock In \emph{Proceedings of the IEEE/CVF Conference on Computer Vision
  and Pattern Recognition}, pages 12976--12985.

\bibitem[{Iashin and Rahtu(2020)}]{iashin2020better}
Vladimir Iashin and Esa Rahtu. 2020.
\newblock A better use of audio-visual cues: Dense video captioning with
  bi-modal transformer.
\newblock \emph{arXiv preprint arXiv:2005.08271}.

\bibitem[{Joo et~al.(2023)Joo, Vo, Yamazaki, and Le}]{joo2023clip}
Hyekang~Kevin Joo, Khoa Vo, Kashu Yamazaki, and Ngan Le. 2023.
\newblock Clip-tsa: Clip-assisted temporal self-attention for weakly-supervised
  video anomaly detection.
\newblock In \emph{2023 IEEE International Conference on Image Processing
  (ICIP)}, pages 3230--3234. IEEE.

\bibitem[{Kay et~al.(2017)Kay, Carreira, Simonyan, Zhang, Hillier,
  Vijayanarasimhan, Viola, Green, Back, Natsev et~al.}]{kay2017kinetics}
Will Kay, Joao Carreira, Karen Simonyan, Brian Zhang, Chloe Hillier, Sudheendra
  Vijayanarasimhan, Fabio Viola, Tim Green, Trevor Back, Paul Natsev, et~al.
  2017.
\newblock The kinetics human action video dataset.
\newblock \emph{arXiv preprint arXiv:1705.06950}.

\bibitem[{Kuehne et~al.(2011)Kuehne, Jhuang, Garrote, Poggio, and
  Serre}]{kuehne2011hmdb}
Hildegard Kuehne, Hueihan Jhuang, Est{\'\i}baliz Garrote, Tomaso Poggio, and
  Thomas Serre. 2011.
\newblock Hmdb: a large video database for human motion recognition.
\newblock In \emph{2011 International conference on computer vision}, pages
  2556--2563. IEEE.

\bibitem[{Lee et~al.(2017)Lee, Huang, Singh, and Yang}]{lee2017unsupervised}
Hsin-Ying Lee, Jia-Bin Huang, Maneesh Singh, and Ming-Hsuan Yang. 2017.
\newblock Unsupervised representation learning by sorting sequences.
\newblock In \emph{Proceedings of the IEEE international conference on computer
  vision}, pages 667--676.

\bibitem[{Li et~al.(2020)Li, Duan, Fang, Gong, and Jiang}]{li2020unicoder}
Gen Li, Nan Duan, Yuejian Fang, Ming Gong, and Daxin Jiang. 2020.
\newblock Unicoder-vl: A universal encoder for vision and language by
  cross-modal pre-training.
\newblock In \emph{Proceedings of the AAAI Conference on Artificial
  Intelligence}, volume~34, pages 11336--11344.

\bibitem[{Li et~al.(2023)Li, Li, Savarese, and Hoi}]{li2023blip}
Junnan Li, Dongxu Li, Silvio Savarese, and Steven Hoi. 2023.
\newblock Blip-2: Bootstrapping language-image pre-training with frozen image
  encoders and large language models.
\newblock \emph{arXiv preprint arXiv:2301.12597}.

\bibitem[{Li et~al.(2022{\natexlab{a}})Li, Li, Xiong, and Hoi}]{li2022blip}
Junnan Li, Dongxu Li, Caiming Xiong, and Steven Hoi. 2022{\natexlab{a}}.
\newblock Blip: Bootstrapping language-image pre-training for unified
  vision-language understanding and generation.
\newblock In \emph{International Conference on Machine Learning}, pages
  12888--12900. PMLR.

\bibitem[{Li et~al.(2021)Li, Selvaraju, Gotmare, Joty, Xiong, and
  Hoi}]{li2021align}
Junnan Li, Ramprasaath Selvaraju, Akhilesh Gotmare, Shafiq Joty, Caiming Xiong,
  and Steven Chu~Hong Hoi. 2021.
\newblock Align before fuse: Vision and language representation learning with
  momentum distillation.
\newblock \emph{Advances in neural information processing systems},
  34:9694--9705.

\bibitem[{Li et~al.(2022{\natexlab{b}})Li, Liu, and Jiao}]{li2022self}
Shuo Li, Fang Liu, and Licheng Jiao. 2022{\natexlab{b}}.
\newblock Self-training multi-sequence learning with transformer for weakly
  supervised video anomaly detection.
\newblock In \emph{Proceedings of the AAAI Conference on Artificial
  Intelligence}, volume~36, pages 1395--1403.

\bibitem[{Li and Wang(2020)}]{li2020learning}
Tianhao Li and Limin Wang. 2020.
\newblock Learning spatiotemporal features via video and text pair
  discrimination.
\newblock \emph{arXiv preprint arXiv:2001.05691}.

\bibitem[{Lu et~al.(2019)Lu, Batra, Parikh, and Lee}]{lu2019vilbert}
Jiasen Lu, Dhruv Batra, Devi Parikh, and Stefan Lee. 2019.
\newblock {ViLBERT}: Pretraining task-agnostic visiolinguistic representations
  for vision-and-language tasks.
\newblock \emph{arXiv preprint arXiv:1908.02265}.

\bibitem[{Man et~al.(2022)Man, Ouyang, Li, Song, and Shao}]{man2022scenario}
Xin Man, Deqiang Ouyang, Xiangpeng Li, Jingkuan Song, and Jie Shao. 2022.
\newblock Scenario-aware recurrent transformer for goal-directed video
  captioning.
\newblock \emph{ACM Transactions on Multimedia Computing, Communications, and
  Applications (TOMM)}, 18(4):1--17.

\bibitem[{Miech et~al.(2020)Miech, Alayrac, Smaira, Laptev, Sivic, and
  Zisserman}]{miech2020end}
Antoine Miech, Jean-Baptiste Alayrac, Lucas Smaira, Ivan Laptev, Josef Sivic,
  and Andrew Zisserman. 2020.
\newblock End-to-end learning of visual representations from uncurated
  instructional videos.
\newblock In \emph{Proceedings of the IEEE/CVF Conference on Computer Vision
  and Pattern Recognition}, pages 9879--9889.

\bibitem[{Miech et~al.(2019)Miech, Zhukov, Alayrac, Tapaswi, Laptev, and
  Sivic}]{miech2019howto100m}
Antoine Miech, Dimitri Zhukov, Jean-Baptiste Alayrac, Makarand Tapaswi, Ivan
  Laptev, and Josef Sivic. 2019.
\newblock Howto100m: Learning a text-video embedding by watching hundred
  million narrated video clips.
\newblock In \emph{Proceedings of the IEEE/CVF International Conference on
  Computer Vision}, pages 2630--2640.

\bibitem[{Ni et~al.(2022)Ni, Peng, Chen, Zhang, Meng, Fu, Xiang, and
  Ling}]{ni2022expanding}
Bolin Ni, Houwen Peng, Minghao Chen, Songyang Zhang, Gaofeng Meng, Jianlong Fu,
  Shiming Xiang, and Haibin Ling. 2022.
\newblock Expanding language-image pretrained models for general video
  recognition.
\newblock In \emph{Computer Vision--ECCV 2022: 17th European Conference, Tel
  Aviv, Israel, October 23--27, 2022, Proceedings, Part IV}, pages 1--18.
  Springer.

\bibitem[{Oord et~al.(2018)Oord, Li, and Vinyals}]{oord2018representation}
Aaron van~den Oord, Yazhe Li, and Oriol Vinyals. 2018.
\newblock Representation learning with contrastive predictive coding.
\newblock \emph{arXiv preprint arXiv:1807.03748}.

\bibitem[{Pan et~al.(2021)Pan, Song, Yang, Jiang, and Liu}]{pan2021videomoco}
Tian Pan, Yibing Song, Tianyu Yang, Wenhao Jiang, and Wei Liu. 2021.
\newblock Videomoco: Contrastive video representation learning with temporally
  adversarial examples.
\newblock In \emph{Proceedings of the IEEE/CVF conference on computer vision
  and pattern recognition}, pages 11205--11214.

\bibitem[{Pang et~al.(2021)Pang, He, Hu, and Li}]{pang2021violence}
Wen-Feng Pang, Qian-Hua He, Yong-jian Hu, and Yan-Xiong Li. 2021.
\newblock Violence detection in videos based on fusing visual and audio
  information.
\newblock In \emph{ICASSP 2021-2021 IEEE international conference on acoustics,
  speech and signal processing (ICASSP)}, pages 2260--2264. IEEE.

\bibitem[{Park et~al.(2023)Park, Kim, Kim, Kim, and Sohn}]{park2023normality}
Seongheon Park, Hanjae Kim, Minsu Kim, Dahye Kim, and Kwanghoon Sohn. 2023.
\newblock Normality guided multiple instance learning for weakly supervised
  video anomaly detection.
\newblock In \emph{Proceedings of the IEEE/CVF Winter Conference on
  Applications of Computer Vision}, pages 2665--2674.

\bibitem[{Patrick et~al.(2020)Patrick, Asano, Kuznetsova, Fong, Henriques,
  Zweig, and Vedaldi}]{patrick2020multi}
Mandela Patrick, Yuki Asano, Polina Kuznetsova, Ruth Fong, Joao~F Henriques,
  Geoffrey Zweig, and Andrea Vedaldi. 2020.
\newblock Multi-modal self-supervision from generalized data transformations.
\newblock In \emph{International Conference on Computer Vision (ICCV)}.

\bibitem[{Piergiovanni et~al.(2020)Piergiovanni, Angelova, and
  Ryoo}]{piergiovanni2020evolving}
AJ~Piergiovanni, Anelia Angelova, and Michael~S Ryoo. 2020.
\newblock Evolving losses for unsupervised video representation learning.
\newblock In \emph{Proceedings of the IEEE/CVF Conference on Computer Vision
  and Pattern Recognition}, pages 133--142.

\bibitem[{Pu and Wu(2022)}]{pu2022audio}
Yujiang Pu and Xiaoyu Wu. 2022.
\newblock Audio-guided attention network for weakly supervised violence
  detection.
\newblock In \emph{2022 2nd International Conference on Consumer Electronics
  and Computer Engineering (ICCECE)}, pages 219--223. IEEE.

\bibitem[{Qian et~al.(2021)Qian, Meng, Gong, Yang, Wang, Belongie, and
  Cui}]{qian2021spatiotemporal}
Rui Qian, Tianjian Meng, Boqing Gong, Ming-Hsuan Yang, Huisheng Wang, Serge
  Belongie, and Yin Cui. 2021.
\newblock Spatiotemporal contrastive video representation learning.
\newblock In \emph{Proceedings of the IEEE/CVF Conference on Computer Vision
  and Pattern Recognition}, pages 6964--6974.

\bibitem[{Radford et~al.(2021)Radford, Kim, Hallacy, Ramesh, Goh, Agarwal,
  Sastry, Askell, Mishkin, Clark et~al.}]{radford2021learning}
Alec Radford, Jong~Wook Kim, Chris Hallacy, Aditya Ramesh, Gabriel Goh,
  Sandhini Agarwal, Girish Sastry, Amanda Askell, Pamela Mishkin, Jack Clark,
  et~al. 2021.
\newblock Learning transferable visual models from natural language
  supervision.
\newblock In \emph{International conference on machine learning}, pages
  8748--8763. PMLR.

\bibitem[{Ramesh et~al.(2021)Ramesh, Pavlov, Goh, Gray, Voss, Radford, Chen,
  and Sutskever}]{ramesh2021zero}
Aditya Ramesh, Mikhail Pavlov, Gabriel Goh, Scott Gray, Chelsea Voss, Alec
  Radford, Mark Chen, and Ilya Sutskever. 2021.
\newblock Zero-shot text-to-image generation.
\newblock In \emph{International Conference on Machine Learning}, pages
  8821--8831. PMLR.

\bibitem[{Sch{\"o}lkopf et~al.(1999)Sch{\"o}lkopf, Williamson, Smola,
  Shawe-Taylor, and Platt}]{scholkopf1999support}
Bernhard Sch{\"o}lkopf, Robert~C Williamson, Alex Smola, John Shawe-Taylor, and
  John Platt. 1999.
\newblock Support vector method for novelty detection.
\newblock \emph{Advances in neural information processing systems}, 12.

\bibitem[{Seo et~al.(2022)Seo, Nagrani, Arnab, and Schmid}]{seo2022end}
Paul~Hongsuck Seo, Arsha Nagrani, Anurag Arnab, and Cordelia Schmid. 2022.
\newblock End-to-end generative pretraining for multimodal video captioning.
\newblock In \emph{Proceedings of the IEEE/CVF Conference on Computer Vision
  and Pattern Recognition}, pages 17959--17968.

\bibitem[{Shafaei et~al.(2021)Shafaei, Smailis, Kakadiaris, and
  Solorio}]{shafaei2021case}
Mahsa Shafaei, Christos Smailis, Ioannis~A Kakadiaris, and Thamar Solorio.
  2021.
\newblock A case study of deep learning based multi-modal methods for
  predicting the age-suitability rating of movie trailers.
\newblock \emph{arXiv preprint arXiv:2101.11704}.

\bibitem[{Soomro et~al.(2012)Soomro, Zamir, and Shah}]{soomro2012ucf101}
Khurram Soomro, Amir~Roshan Zamir, and Mubarak Shah. 2012.
\newblock Ucf101: A dataset of 101 human actions classes from videos in the
  wild.
\newblock \emph{arXiv preprint arXiv:1212.0402}.

\bibitem[{Strasburger(1989)}]{strasburger1989adolescent}
V.~C. Strasburger. 1989.
\newblock Adolescent sexuality and the media.
\newblock \emph{Pediatric Clinics of North America}, 36(3):747--773.

\bibitem[{Sun et~al.(2022)Sun, Ke, Rahmani, Bennamoun, Wang, and
  Liu}]{sun2022human}
Zehua Sun, Qiuhong Ke, Hossein Rahmani, Mohammed Bennamoun, Gang Wang, and Jun
  Liu. 2022.
\newblock Human action recognition from various data modalities: A review.
\newblock \emph{IEEE transactions on pattern analysis and machine
  intelligence}.

\bibitem[{Tan and Bansal(2019)}]{tan2019lxmert}
Hao Tan and Mohit Bansal. 2019.
\newblock Lxmert: Learning cross-modality encoder representations from
  transformers.
\newblock \emph{arXiv preprint arXiv:1908.07490}.

\bibitem[{Tian et~al.(2021)Tian, Pang, Chen, Singh, Verjans, and
  Carneiro}]{tian2021weakly}
Yu~Tian, Guansong Pang, Yuanhong Chen, Rajvinder Singh, Johan~W Verjans, and
  Gustavo Carneiro. 2021.
\newblock Weakly-supervised video anomaly detection with robust temporal
  feature magnitude learning.
\newblock In \emph{Proceedings of the IEEE/CVF international conference on
  computer vision}, pages 4975--4986.

\bibitem[{Tong et~al.(2022)Tong, Song, Wang, and Wang}]{tong2022videomae}
Zhan Tong, Yibing Song, Jue Wang, and Limin Wang. 2022.
\newblock Videomae: Masked autoencoders are data-efficient learners for
  self-supervised video pre-training.
\newblock \emph{Advances in neural information processing systems},
  35:10078--10093.

\bibitem[{Udandarao et~al.(2020)Udandarao, Maiti, Srivatsav, Vyalla, Yin, and
  Shah}]{udandarao2020cobra}
Vishaal Udandarao, Abhishek Maiti, Deepak Srivatsav, Suryatej~Reddy Vyalla,
  Yifang Yin, and Rajiv~Ratn Shah. 2020.
\newblock Cobra: Contrastive bi-modal representation algorithm.
\newblock \emph{arXiv preprint arXiv:2005.03687}.

\bibitem[{Wang et~al.(2020)Wang, Jiao, and Liu}]{wang2020self}
Jiangliu Wang, Jianbo Jiao, and Yun-Hui Liu. 2020.
\newblock Self-supervised video representation learning by pace prediction.
\newblock In \emph{Computer Vision--ECCV 2020: 16th European Conference,
  Glasgow, UK, August 23--28, 2020, Proceedings, Part XVII 16}, pages 504--521.
  Springer.

\bibitem[{Wang et~al.(2023)Wang, Huang, Zhao, Tong, He, Wang, Wang, and
  Qiao}]{wang2023videomae}
Limin Wang, Bingkun Huang, Zhiyu Zhao, Zhan Tong, Yinan He, Yi~Wang, Yali Wang,
  and Yu~Qiao. 2023.
\newblock Videomae v2: Scaling video masked autoencoders with dual masking.
\newblock In \emph{Proceedings of the IEEE/CVF Conference on Computer Vision
  and Pattern Recognition}, pages 14549--14560.

\bibitem[{Wilson(2008)}]{wilson2008media}
Barbara~J Wilson. 2008.
\newblock Media and children's aggression, fear, and altruism.
\newblock \emph{The future of children}, pages 87--118.

\bibitem[{Wu et~al.(2022{\natexlab{a}})Wu, Hsieh, Chen, Fuh, and
  Liu}]{wu2022self}
Jhih-Ciang Wu, He-Yen Hsieh, Ding-Jie Chen, Chiou-Shann Fuh, and Tyng-Luh Liu.
  2022{\natexlab{a}}.
\newblock Self-supervised sparse representation for video anomaly detection.
\newblock In \emph{European Conference on Computer Vision}, pages 729--745.
  Springer.

\bibitem[{Wu and Liu(2021)}]{wu2021learning}
Peng Wu and Jing Liu. 2021.
\newblock Learning causal temporal relation and feature discrimination for
  anomaly detection.
\newblock \emph{IEEE Transactions on Image Processing}, 30:3513--3527.

\bibitem[{Wu et~al.(2020)Wu, Liu, Shi, Sun, Shao, Wu, and Yang}]{wu2020not}
Peng Wu, Jing Liu, Yujia Shi, Yujia Sun, Fangtao Shao, Zhaoyang Wu, and Zhiwei
  Yang. 2020.
\newblock Not only look, but also listen: Learning multimodal violence
  detection under weak supervision.
\newblock In \emph{Computer Vision--ECCV 2020: 16th European Conference,
  Glasgow, UK, August 23--28, 2020, Proceedings, Part XXX 16}, pages 322--339.
  Springer.

\bibitem[{Wu et~al.(2022{\natexlab{b}})Wu, Liu, and Liu}]{wu2022weakly}
Peng Wu, Xiaotao Liu, and Jing Liu. 2022{\natexlab{b}}.
\newblock Weakly supervised audio-visual violence detection.
\newblock \emph{IEEE Transactions on Multimedia}.

\bibitem[{Wu et~al.(2023)Wu, Zhou, Pang, Zhou, Yan, Wang, and
  Zhang}]{wu2023vadclip}
Peng Wu, Xuerong Zhou, Guansong Pang, Lingru Zhou, Qingsen Yan, Peng Wang, and
  Yanning Zhang. 2023.
\newblock Vadclip: Adapting vision-language models for weakly supervised video
  anomaly detection.
\newblock \emph{arXiv preprint arXiv:2308.11681}.

\bibitem[{Xu et~al.(2019)Xu, Xiao, Zhao, Shao, Xie, and Zhuang}]{xu2019self}
Dejing Xu, Jun Xiao, Zhou Zhao, Jian Shao, Di~Xie, and Yueting Zhuang. 2019.
\newblock Self-supervised spatiotemporal learning via video clip order
  prediction.
\newblock In \emph{Proceedings of the IEEE/CVF Conference on Computer Vision
  and Pattern Recognition}, pages 10334--10343.

\bibitem[{Xu et~al.(2022)Xu, Zhu, and Clifton}]{xu2022multimodal}
Peng Xu, Xiatian Zhu, and David~A Clifton. 2022.
\newblock Multimodal learning with transformers: A survey.
\newblock \emph{arXiv preprint arXiv:2206.06488}.

\bibitem[{Yang et~al.(2020)Yang, Xu, Dai, and Zhou}]{yang2020video}
Ceyuan Yang, Yinghao Xu, Bo~Dai, and Bolei Zhou. 2020.
\newblock Video representation learning with visual tempo consistency.
\newblock \emph{arXiv preprint arXiv:2006.15489}.

\bibitem[{Yang et~al.(2023{\natexlab{a}})Yang, Yang, Li, Luo, Jiang, Zhang, and
  Wang}]{yang2023bert}
Jianxi Yang, Xiaoxia Yang, Ren Li, Mengting Luo, Shixin Jiang, Yue Zhang, and
  Di~Wang. 2023{\natexlab{a}}.
\newblock Bert and hierarchical cross attention-based question answering over
  bridge inspection knowledge graph.
\newblock \emph{Expert Systems with Applications}, page 120896.

\bibitem[{Yang et~al.(2023{\natexlab{b}})Yang, Nakashima, and
  Takemura}]{yang2023multi}
Zekun Yang, Yuta Nakashima, and Haruo Takemura. 2023{\natexlab{b}}.
\newblock Multi-modal humor segment prediction in video.
\newblock \emph{Multimedia Systems}, pages 1--10.

\bibitem[{Yu et~al.(2022)Yu, Liu, Cheng, Feng, and Zhang}]{yu2022modality}
Jiashuo Yu, Jinyu Liu, Ying Cheng, Rui Feng, and Yuejie Zhang. 2022.
\newblock Modality-aware contrastive instance learning with self-distillation
  for weakly-supervised audio-visual violence detection.
\newblock In \emph{Proceedings of the 30th ACM International Conference on
  Multimedia}, pages 6278--6287.

\bibitem[{Yuan et~al.(2021)Yuan, Lin, Kuen, Zhang, Wang, Maire, Kale, and
  Faieta}]{yuan2021multimodal}
Xin Yuan, Zhe Lin, Jason Kuen, Jianming Zhang, Yilin Wang, Michael Maire,
  Ajinkya Kale, and Baldo Faieta. 2021.
\newblock Multimodal contrastive training for visual representation learning.
\newblock In \emph{Proceedings of the IEEE/CVF Conference on Computer Vision
  and Pattern Recognition}, pages 6995--7004.

\bibitem[{Zhang et~al.(2023{\natexlab{a}})Zhang, Li, Qi, Wang, Qing, Huang, and
  Yang}]{zhang2023exploiting}
Chen Zhang, Guorong Li, Yuankai Qi, Shuhui Wang, Laiyun Qing, Qingming Huang,
  and Ming-Hsuan Yang. 2023{\natexlab{a}}.
\newblock Exploiting completeness and uncertainty of pseudo labels for weakly
  supervised video anomaly detection.
\newblock In \emph{Proceedings of the IEEE/CVF Conference on Computer Vision
  and Pattern Recognition}, pages 16271--16280.

\bibitem[{Zhang et~al.(2022)Zhang, Zhang, and Pan}]{zhang2022hierarchical}
Litian Zhang, Xiaoming Zhang, and Junshu Pan. 2022.
\newblock Hierarchical cross-modality semantic correlation learning model for
  multimodal summarization.
\newblock In \emph{Proceedings of the AAAI Conference on Artificial
  Intelligence}, volume~36, pages 11676--11684.

\bibitem[{Zhang et~al.(2021)Zhang, Shafaei, Gonzalez, and
  Solorio}]{zhang2021none}
Yigeng Zhang, Mahsa Shafaei, Fabio Gonzalez, and Thamar Solorio. 2021.
\newblock From none to severe: {P}redicting severity in movie scripts.
\newblock In \emph{Findings of the Association for Computational Linguistics:
  EMNLP 2021}, pages 3951--3956, Punta Cana, Dominican Republic. Association
  for Computational Linguistics.

\bibitem[{Zhang et~al.(2023{\natexlab{b}})Zhang, Shafaei, Gonzalez, and
  Solorio}]{zhang2023positive}
Yigeng Zhang, Mahsa Shafaei, Fabio Gonzalez, and Thamar Solorio.
  2023{\natexlab{b}}.
\newblock Positive and risky message assessment for music products.
\newblock \emph{arXiv preprint arXiv:2309.10182}.

\bibitem[{Zheng et~al.(2020)Zheng, Guo, and Kordjamshidi}]{zheng2020cross}
Chen Zheng, Quan Guo, and Parisa Kordjamshidi. 2020.
\newblock Cross-modality relevance for reasoning on language and vision.
\newblock \emph{arXiv preprint arXiv:2005.06035}.

\bibitem[{Zhou et~al.(2023)Zhou, Yu, and Yang}]{zhou2023dual}
Hang Zhou, Junqing Yu, and Wei Yang. 2023.
\newblock Dual memory units with uncertainty regulation for weakly supervised
  video anomaly detection.
\newblock \emph{arXiv preprint arXiv:2302.05160}.

\bibitem[{Zhu et~al.(2022)Zhu, Zhou, Zhang, Liu, Jiao, Zhang, Dai, Jiang, Li,
  and Wei}]{zhu2022vatlm}
Qiushi Zhu, Long Zhou, Ziqiang Zhang, Shujie Liu, Binxing Jiao, Jie Zhang,
  Lirong Dai, Daxin Jiang, Jinyu Li, and Furu Wei. 2022.
\newblock Vatlm: Visual-audio-text pre-training with unified masked prediction
  for speech representation learning.
\newblock \emph{arXiv preprint arXiv:2211.11275}.

\end{thebibliography}

\section{Appendix}
\label{sec:appendix}
The appendix provides detailed explanations about pretraining approaches (\ref{pretraining-approaches}), pretraining datasets (\ref{pre-dataset}) and the experimental setup (\ref{exp-setup}) for both pretraining and fine-tuning processes.

\subsection{Pretraining Approaches}
\label{pretraining-approaches}
In this section, we explain the pretraining approaches that we considered to warm-start our multimodal model. 

\subsubsection{Video-Audio-Text Matching}
\label{pretraining-matching}

In our model, we pretrain for the tasks of Video-Text Matching (VTM), Video-Audio Matching (VAM), and Audio-Text Matching (ATM). We utilize three separate Multi-Layer Perceptron (MLP) blocks, for predicting VTM, ATM, and VAM. For each task, the corresponding representations of the two modalities are combined and processed through the MLP block. For instance, to predict ATM, the text and audio representations are concatenated and passed through the MLP block. A softmax function is employed to generate an output of either 0 or 1, where 0 signifies a mismatch between the modalities and 1 indicates a match.

In the pretraining process, HICCAP reads a batch of video/audio/text triplets from the dataset at each iteration. At this stage, all modalities are matched together, resulting in a matching label of 1. However, in order to create mismatch triplets, we employed a method of randomly generating mismatches among the batch samples. 
This method takes the batch of video, audio, and text vector triplets as input. Then for each sample (triplet) of a batch, it randomly selects one of the three modalities, and with a probability of $p$ replaces it with the same modality from another sample in the batch. 
In case of replacement, the matching label of the modality and the other modalities would change to zero.
As a result, when the selected modality is replaced, the label for the matching between that modality and the other modalities is set to zero. For example, if the video modality is chosen and then replaced with another video sample, the labels for VTM and VAM for the current sample would be 0, but the label for ATM would remain 1. 
This process is repeated for all samples in the batch, resulting in a new batch of samples with updated labels. During the training process, this function is called each time, and its output is used as the input for the Video-Text Matching, Video-Audio Matching, and Audio-Text Matching fully connected layers.
During the pretraining, we compute the loss for Video-Text Matching (VTM), Video-Audio Matching (VAM), and Audio-Text Matching (ATM) tasks and aggregate all losses together to find the final loss. 
\begin{equation}
\resizebox{0.8\hsize}{!}{$\begin{array}{ll}
\mathcal{L} = \lambda_{VTM}\mathcal{L}_{VTM} + \lambda_{VAM}\mathcal{L}_{VAM} + \lambda_{ATM}\mathcal{L}_{ATM}\end{array}$}
\end{equation}


\subsubsection{Multimodal Contrastive Learning}

Inspired by \cite{akbari2021vatt,gurram2022lava}, we utilize common space projection in conjunction with contrastive learning to effectively train our model. Specifically, we leverage the video, audio, and text representation outputs generated from the hierarchical-cross-attention part to establish a semantically common space mapping. We represent the embedding for modality $m$ as $z_m$. 
For this purpose, we define the projection function $g_{m,m'}$ to project both $z_m$ and $z_{m'}$ into a multi-modal latent space $m$ and $m'$. Within these multi-modal latent spaces, we utilize a contrastive framework to compare embeddings using a cosine similarity function. Consequently, for all $i$, when $m'\neq m$, we achieve high similarity between $g_{m,m'}(z_{m,i})$ and $g_{m,m'}(z_{m',i})$, and for all $i \neq j$ when $m' \neq m$, we obtain low similarity $g_{m,m'}(z_{m,i})$ and $g_{m,m'}(z_{m',j})$. 
We use a linear projection (three linear layers with ReLU and batch normalization) for the $g_{m,m'}$ mapping.

\textbf{Cross-Modal Contrastive Loss: }
We illustrate the positive training pairs as various modalities from the same sample, while negative training pairs are generated from different modalities of distinct samples within a batch. Subsequently, a minibatch containing N video samples resulted in $N$ positive pairs and $N^2 - N$ negative pairs. Utilizing Noise Contrastive Estimation (NCE) \cite{gutmann2010noise} as our contrastive loss, we aim to enhance the similarity between positive pairs and increase the dissimilarity between negative pairs in the associated joint embedding space.
The NCE loss is expressed as follows (omitting the projection functions for simplicity):
\begin{multline}
\resizebox{0.85\hsize}{!}{$\begin{array}{ll}
NCE(z_m, z_m' ) = - \log( \\
\dfrac{\Sigma^{N}_{i=0}\exp(z^{T}_{m,i}z^{ }_{m',i}/\tau)}{\Sigma^{N}_{i=0}\exp(z^{T}_{m,i}z^{ }_{m',i}/\tau)+\Sigma^{N}_{i=0}\Sigma^{N}_{j\neq i}\exp(z^{T}_{m,i}z^{ }_{m',j}/\tau)})\end{array}$}
\end{multline}

In accordance with \cite{chen2021multimodal}, we apply NCE to audio-video, audio-text, and video-text pairs:
\begin{equation}
\resizebox{0.7\hsize}{!}{$\begin{array}{ll}
\mathcal{L}_{AV}(z_a, z_v) = NCE(g_{av}(z_a), g_{av}(z_v))\end{array}$}
\end{equation}
\begin{equation}
\resizebox{0.7\hsize}{!}{$\begin{array}{ll}
\mathcal{L}_{AT}(z_a, z_t) = NCE(g_{at}(z_a), g_{at}(z_t))\end{array}$}
\end{equation}
\begin{equation}
\resizebox{0.7\hsize}{!}{$\begin{array}{ll}
\mathcal{L}_{VT}(z_v, z_t) = NCE(g_{vt}(z_v), g_{vt}(z_t))\end{array}$}
\end{equation}

By aggregating all losses, we establish the complete loss for this phase of pretraining:
\begin{equation}
\resizebox{0.7\hsize}{!}{$\begin{array}{ll}
\mathcal{L} = \lambda_{A,V}\mathcal{L}_{AV} + \lambda_{A,T}\mathcal{L}_{AT} + \lambda_{V,T}\mathcal{L}_{VT}\end{array}$}
\end{equation}
where $\lambda_{m,m'}$ corresponds to the weight for the modality pair $m$ and $m'$ that is a learnable parameter and will be tuned during training.

\subsection{Pretraining Datasets}
\label{pre-dataset}

We pretrain our model on various datasets containing video clips with corresponding audio clips and descriptions. To increase the versatility of our model, we use a diverse set of domains, such as human actions, movies, and personal collections.
For this purpose, we use \textbf{Kinetics-400} dataset ~\cite{kay2017kinetics} and part of \textbf{HowTo100M} dataset~\cite{miech2019howto100m}, each with a different distribution of video, audio, and description.

\textbf{Kinetics-400 Dataset ~\cite{kay2017kinetics}} includes 400 diverse human action classes, each featuring at least 400 video clips. These clips, which have a duration of around 10 seconds, are extracted from individual YouTube videos. The variety of action categories spans from human-object interactions, such as playing musical instruments, to human-human interactions like shaking hands.

\textbf{HowTo100M Dataset~\cite{miech2019howto100m}} is a huge collection of YouTube videos focused on instructional videos where content producers teach complicated activities with the specific goal of demonstrating the visual information displayed on the screen. It has 136M video clips with captions which include 23K activities from various domains extracted from 1.2M YouTube videos. Each video has a narration accessible as automatically downloaded subtitles from YouTube.


\subsection{Experimental Setup}
\label{exp-setup}

In this study, we employed the AdamW optimizer with a weight decay of 0.02, betas set at the default values of (0.9,0.999), and eps of 1e-8. Additionally, we used Pytorch's adaptive scheduler, which adjusts the learning rate by a factor of 0.5 and maintains a minimum learning rate of 1e-8.
Our experiments were conducted on an A100 GPU with 40GB of memory, allowing for a batch size of 16. We performed 30 epochs, saving the optimal model for further use.
Utilizing the A100 GPU and a batch size of 16, each pretraining epoch took approximately 2960 seconds (2630 seconds for training and 330 seconds for validation). In the fine-tuning phase, each epoch required around 250 seconds (220 seconds for training and 30 seconds for validation). Our implementation utilized Python 3.11.1, scikit-learn 1.2, and torchvision 0.14.1, along with NVIDIA-SMI 515.48.07, Driver Version 515.48.07, and CUDA Version 11.7.


\end{document}



\section{Appendix}
The appendix provides detailed explanations about datasets~\ref{pretraining-appendix} and the experimental setup \ref{experimental-appendix} for both pretraining and fine-tuning processes.

\subsection{Pretraining Datasets}
\label{pretraining-appendix}
We pretrain our model on various datasets containing video clips with corresponding audio clips and descriptions. To increase the versatility of our model, we use a diverse set of domains, such as human actions, movies, and personal collections.
For this purpose, we use \textbf{Kinetics-400} dataset [31] and part of two other datasets, \textbf{HowTo100M} dataset [37] and \textbf{Violent Scenes} dataset [16], each with a different distribution of video, audio, and description.

\textbf{Kinetics-400 Dataset [31]} includes 400 diverse human action classes, each featuring at least 400 video clips. These clips, which have a duration of around 10 seconds, are extracted from individual YouTube videos. The variety of action categories spans from human-object interactions, such as playing musical instruments, to human-human interactions like shaking hands.

\textbf{HowTo100M Dataset [37]} is a huge collection of YouTube videos focused on instructional videos where content producers teach complicated activities with the specific goal of demonstrating the visual information displayed on the screen. It has 136M video clips with captions which include 23K activities from various domains extracted from 1.2M YouTube videos. Each video has a narration accessible as automatically downloaded subtitles from YouTube.

\textbf{Violent Scenes Dataset [16]} is a collection of 32 movies from a variety of genres (including both violent and non-violent movies) and is based on the extraction of violent events from movies and online videos. Violent parts and high-level movie concepts were annotated at 25 frames per second. In the ground-truth files, only the parts related to the targeted events were marked and shown.

\subsection{Experimental Setup}
\label{experimental-appendix}

In this study, we employed the AdamW optimizer with a weight decay of 0.02, betas set at the default values of (0.9,0.999), and eps of 1e-8. Additionally, we used Pytorch's adaptive scheduler, which adjusts the learning rate by a factor of 0.5 and maintains a minimum learning rate of 1e-8.
Our experiments were conducted on an A100 GPU with 40GB of memory, allowing for a batch size of 16. We performed 30 epochs, saving the optimal model for further use (the model with the lowest validation loss during pretraining and the highest average accuracy during fine-tuning).
Utilizing the A100 GPU and a batch size of 16, each pretraining epoch took approximately 2960 seconds (2630 seconds for training and 330 seconds for validation). In the fine-tuning phase, each epoch required around 250 seconds (220 seconds for training and 30 seconds for validation). Our implementation utilized Python 3.11.1, scikit-learn 1.2, and torchvision 0.14.1, along with NVIDIA-SMI 515.48.07, Driver Version 515.48.07, and CUDA Version 11.7.